\pdfoutput=1
\documentclass[11pt]{article}
\usepackage[final]{acl}

\usepackage{times}
\usepackage{latexsym}
\usepackage[T1]{fontenc}
\usepackage[utf8]{inputenc}
\usepackage{microtype}
\usepackage{inconsolata}
\usepackage{graphicx}
\usepackage{amsfonts}
\usepackage{amsmath, amssymb, mathtools}
\usepackage{tikz}
\usepackage{bm}
\usepackage{bold-extra}
\usepackage{subcaption}
\usepackage{tikz-dependency}
\usepackage{algorithm}
\usepackage[noend]{algpseudocode}
\usepackage{multirow}
\usepackage{slashbox}

\definecolor{bcolor}{HTML}{0000FF}
\definecolor{sbcolor}{HTML}{E62020}
\definecolor{tcolor}{HTML}{018E42}

\newcommand{\tor}{\textcolor{tcolor}{\texttt{/}}}
\newcommand{\tcr}{\textcolor{tcolor}{\texttt{>}}}
\newcommand{\tol}{\textcolor{tcolor}{\texttt{<}}}
\newcommand{\tcl}{\textcolor{tcolor}{\texttt{\textbackslash}}}

\newcommand{\bor}{\textcolor{bcolor}{\texttt{/}}}
\newcommand{\bcr}{\textcolor{bcolor}{\texttt{>}}}
\newcommand{\bol}{\textcolor{bcolor}{\texttt{<}}}
\newcommand{\bcl}{\textcolor{bcolor}{\texttt{\textbackslash}}}

\newcommand{\bori}[1]{\textcolor{bcolor}{\texttt{/}}\ensuremath{_{\textcolor{bcolor}{\text{#1}}}}}
\newcommand{\bcri}[1]{\textcolor{bcolor}{\texttt{>}}\ensuremath{_{\textcolor{bcolor}{\text{#1}}}}}
\newcommand{\boli}[1]{\textcolor{bcolor}{\texttt{<}}\ensuremath{_{\textcolor{bcolor}{\text{#1}}}}}

\newcommand{\sor}{\textcolor{sbcolor}{\textbf{\texttt{/}}}}
\newcommand{\scr}{\textcolor{sbcolor}{\textbf{\texttt{>}}}}
\newcommand{\sol}{\textcolor{sbcolor}{\textbf{\texttt{<}}}}
\newcommand{\scl}{\textcolor{sbcolor}{\textbf{\texttt{\textbackslash}}}}

\newcommand{\scri}[1]{\textcolor{sbcolor}{\textbf{\texttt{>}}}\ensuremath{_{\textcolor{sbcolor}{\text{\textbf{#1}}}}}}

\def\SPSB#1#2{\rlap{\textsuperscript{#1}}\SB{#2}}

\def\SB#1{\textsubscript{#1}}

\title{Hierarchical Bracketing Encodings for Dependency Parsing as Tagging}

\author{
    Ana Ezquerro\textsuperscript{1} \\ \texttt{ana.ezquerro} \And 
    \hspace{-3em} David Vilares\textsuperscript{1} \\ \hspace{-3em}\texttt{david.vilares} \And 
    \hspace{-4em}Anssi Yli-Jyrä\textsuperscript{2,3} \\ \hspace{-4.5em}\texttt{anssi.yli-jyra} \And 
    \hspace{-3em}Carlos Gómez-Rodríguez\textsuperscript{1} \\ \hspace{-3em}\texttt{carlos.gomez} \\ \AND\\[-1cm]
    \textsuperscript{1}Universidade da Coruña, CITIC (\texttt{@udc.es})\\
    \textsuperscript{2}Tampere University, Math. Res. Centre, Computing Sciences (\texttt{@tuni.fi}) \\
    \textsuperscript{3}University of Helsinki, Faculty of Arts, Digital Humanities (\texttt{@helsinki.fi}) \\
  }

\definecolor{onp}{HTML}{DDBDF5}
\definecolor{op}{HTML}{67B7F5}
\definecolor{dm}{HTML}{F55533}
\definecolor{b4}{HTML}{F4D054}
\definecolor{b7}{HTML}{94F5A2}
\definecolor{h}{HTML}{fc8d62}
\usepackage{tikz}
\newcommand{\cbullet}[1]{
\hspace{-0.4em}
\begin{tikzpicture}
\filldraw[fill=#1,draw=black] circle (3pt);
\end{tikzpicture}
\hspace{-0.4em}
}
\newcommand{\cdiamond}[1]{
\hspace{-0.4em}
\begin{tikzpicture}
\filldraw[fill=#1,draw=black, rotate=45] rectangle (5pt,5pt);
\end{tikzpicture}
\hspace{-0.5em}
}
\newcommand{\csquare}[1]{
\hspace{-0.3em}
\begin{tikzpicture}
\filldraw[fill=#1,draw=black] rectangle (7pt,7pt);
\end{tikzpicture}
\hspace{-0.5em}
}

\begin{document}
\maketitle
\begin{abstract}
We present a family of encodings for sequence labeling dependency parsing, based on the concept of hierarchical bracketing. We prove that the existing 4-bit projective encoding belongs to this family, but it is suboptimal in the number of labels used to encode a tree. We derive an optimal hierarchical bracketing, which minimizes the number of symbols used and encodes projective trees using only 12 distinct labels (vs. 16 for the 4-bit encoding). We also extend optimal hierarchical bracketing to support arbitrary non-projectivity in a more compact way than previous encodings. Our new encodings yield competitive accuracy on a diverse set of treebanks.
\end{abstract}

\section{Introduction}
Sequence labeling with contextualized representations enables efficient dependency parsers that function as taggers while achieving high or even state-of-the-art performance~\citep{amini-etal-2023-hexatagging,gomez-rodriguez-etal-2023-4}. This paradigm has two key aspects: a powerful sequence encoder—often a pre-trained model—and a linearization algorithm that transforms dependency trees into label sequences with a strict one-to-one mapping. Regarding linearizations, various strategies have been proposed. Early approaches represented a word’s head as an offset~\citep{strzyz-etal-2019-viable,lacroix2019dependency}, which could be absolute, relative to the dependent, or based on word properties. Another strategy is bracketing encodings~\citep{yli-jyra-gomez-rodriguez-2017-generic,strzyz-etal-2019-viable}, where each word label encodes a set of incoming or outgoing arcs, unlike positional encodings that explicitly encode the head. A third approach, transition-based encodings \cite{gomez-rodriguez-etal-2020-unifying}, assigns labels to transition subsequences split by read (shift-like) transitions.

These strategies suffer from an unbounded label space. Instead, recent work proposes fixed-space solutions. \citet{amini-etal-2023-hexatagging} proposed hexatagging, a projective tree linearization with eight labels but requiring an intermediate representation. \citet{gomez-rodriguez-etal-2023-4} introduced 4- and 7-bit labels for projective and 2-planar trees, with bounded label spaces of 16 and 128 labels.

This work introduces a new family of hierarchical bracketing encodings, based on rope cover~\cite{yli-jyra-2019-transition}. We show that the existing 4-bit projective encoding is part of this family, but suboptimal in terms of compactness. We introduce and test its optimal counterpart, and show that both bracketings improve accuracy for non-projective trees when the pseudo-projective transformation is applied. We also generalize the optimal bracketing to directly encode arbitrary non-projective trees.

\section{Preliminaries}

Let $w_1 \ldots w_n$ be an input string. A dependency graph is a directed graph $G=(V,E)$ where $V=\{0, \ldots, n\}$. An edge $(i,j)$ in $E$ is called a dependency from word $w_i$ to $w_j$. $w_i$ is the head or parent, $w_j$ is the dependent. Index $0$ is used for a dummy node used when we deal with trees. We use the notation $i \rightarrow j$ to specify that such a dependency is a rightward arc (i.e., when $i<j$) and $i \leftarrow j$ when it is a leftward arc ($i>j$). When direction is not specified, we use $(i,j)$. Node $k$ is a descendant of node $i$ if there is a path from $i$ to $k$.

An arc $(i,j)$ is said to \emph{cross} the arc $(k,l)$ if $\min(i,j) < \min(k,l) < \max(i,j) < \max(k,l)$ or $\min(k,l) < \min(i,j) < \max(k,l) < \max(i,j)$.  An arc $(i,j)$ is said to \emph{cover} arc $(k,l)$ if $\min(i,j) \le \min(k,l) < \max(k,l) \le \max(i,j)$. 

A dependency graph is a tree if it has no cycles and every node has exactly one parent except for the dummy node $0$, which has no parent. A tree is projective if it has no crossing arcs. An equivalent definition (see \citet{nivre-2006-constraints}) will be more directly useful in some proofs: a dependency tree is projective if, for each arc $(i,j)$, all nodes located between $i$ and $j$ are descendants of either $i$ or $j$.
 
\section{Brackets and superbrackets}

\subsection{Bracketing encodings}

\begin{figure*}[tbp]
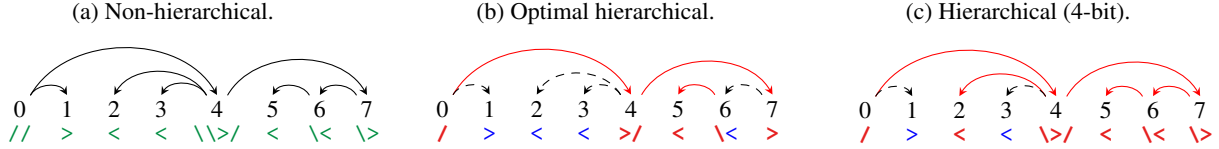
\centering\small 
    \begin{subfigure}[t]{0.3\textwidth}\centering\footnotesize
        \caption{\label{subfig:naive}Non-hierarchical.}
        \vspace{0.1em}
        \begin{dependency}[arc edge,hide label]
           \begin{deptext}[column sep=2pt,minimum width=0.55cm]
              0 \& 1 \& 2 \& 3 \& 4 \& 5 \& 6 \& 7 \\
              \tor \tor \& \tcr \& \tol \& \tol \& \tcl \tcl \tcr \tor \& \tol \& \tcl \tol \& \tcl \tcr\\
           \end{deptext}
           \depedge{1}{5}{}
           \depedge{1}{2}{}
           \depedge{5}{3}{}
           \depedge{5}{4}{}
           \depedge{5}{8}{}
           \depedge{8}{7}{}
           \depedge{7}{6}{}
        \end{dependency}
    \end{subfigure}
    \hfill
    \begin{subfigure}[t]{0.3\textwidth}\centering\footnotesize
        \caption{\label{subfig:optimal}Optimal hierarchical.}
        \vspace{0.2em}
        \begin{dependency}[arc edge,hide label]
           \begin{deptext}[column sep=2pt,minimum width=0.55cm]
              0 \& 1 \& 2 \& 3 \& 4 \& 5 \& 6 \& 7 \\
              \sor \& \bcr \& \bol \& \bol \& \scr \sor \& \sol \& \scl\bol \& \scr \\
           \end{deptext}
           \depedge[edge style={red}]{1}{5}{}
           \depedge[edge style={dashed}]{5}{3}{}
           \depedge[edge style={dashed}]{1}{2}{}
           \depedge[edge style={dashed}]{5}{4}{}
           \depedge[edge style={red}]{5}{8}{}
           \depedge[edge style={dashed}]{8}{7}{}
           \depedge[edge style={red}]{7}{6}{}
        \end{dependency}
    \end{subfigure}
    \hfill 
    \begin{subfigure}[t]{0.3\textwidth}\centering\footnotesize
        \caption{\label{subfig:hierarchical}Hierarchical (4-bit).}
        \begin{dependency}[arc edge,hide label, label style={above}]
           \begin{deptext}[column sep=2pt,minimum width=0.55cm]
              0 \& 1 \& 2 \& 3 \& 4 \& 5 \& 6 \& 7 \\
              \sor \& \bcr \& \sol \& \bol \& \scl \scr \sor \& \sol \& \scl\sol \& \scl\scr \\
           \end{deptext}
           \depedge[edge style={red}]{1}{5}{}
           \depedge[edge style={red}, label style={inner sep=.1ex}]{5}{3}{\textcolor{red}{$\ast$}}
           \depedge[edge style={dashed}]{1}{2}{}
           \depedge[edge style={dashed}]{5}{4}{}
           \depedge[edge style={red}]{5}{8}{}
           \depedge[edge style={red}]{8}{7}{}
           \depedge[edge style={red}]{7}{6}{}
        \end{dependency}
    \end{subfigure}
    \vspace{-0.5em}
    \caption{\label{fig:simpletree}Example showing the same tree encoded with the standard (non-hierarchical) bracketing encoding (Figure \ref{subfig:naive}), and hierarchical bracketing encodings (Figures \ref{subfig:optimal} and \ref{subfig:hierarchical}). In the latter, structural arcs are shown in red and solid, whereas auxiliary arcs are black and dashed. Figure \ref{subfig:optimal} corresponds to the optimal hierarchical bracketing where $0 \rightarrow 4$, $4\to 7$ and $6\to 5$ are the structural arcs. Figure \ref{subfig:hierarchical} corresponds to the 4-bit encoding.
    }
\end{figure*}

Bracketing encodings for dependency parsing are based on the idea of representing dependencies using symbols that behave as balanced brackets. In the earliest and most basic bracketing encoding \citep{strzyz-etal-2019-viable,strzyz-etal-2020-bracketing}, a right arc from a word $w_i$ to $w_j$ is encoded by including an opening bracket {\tor} in the label of $w_i$, which matches the corresponding closing bracket {\tcr} in the label of $w_j$. Similarly, a left arc from $w_j$ to $w_i$ is represented by an opening bracket {\tol} in the label of $w_i$ that matches a closing bracket {\tcl} in the label of $w_j$. The result is shown in Figure~\ref{subfig:naive}, where the label associated with each given word contains one {\tol} (resp. {\tcr}) symbol per incoming arc from the right (left) and one {\tor} (resp. {\tcl}) symbol per outgoing arc towards the right (left). To decode the brackets back into a tree, we read them from left to right, using a stack. When reading an opening bracket ({\tor} or {\tol}) in the label of a word $w_i$, we push it to the stack, associating it with the index $i$. When reading a closing bracket {\tcr} in the label of $w_j$, we pop the matching bracket {\tor} from the top of the stack, check its associated index $i$, and create the right arc from $w_i$ to $w_j$. Finally, when reading a closing bracket {\tcl}, we proceed analogously, popping a {\tol} from the top of the stack and creating a left arc. This decoding method will recover the original tree if it is \emph{projective} (i.e., without crossing arcs), as the nested matching of brackets prevents them from being paired in any way that would create crossing arcs.\footnote{The original implementation in~\citep{strzyz-etal-2019-viable,strzyz-etal-2020-bracketing} does not decode like this. It decodes using two separate stacks for left and right arcs, and then the coverage is extended to support crossing arcs in opposite directions. Here we assume projective decoding with a single stack for didactic reasons, as it helps us establish the relation to our novel encodings. Extensions for non-projectivity will be presented later.} Extensions to support crossing arcs exist~\citep{strzyz-etal-2020-bracketing}, but we will stick to projective trees in this section.

This encoding obtains good practical results, especially in low-resource setups~\citep{munoz-ortiz-etal-2021-linearizations}. Yet, it has the disadvantage that it is \emph{unbounded} in terms of label set size, as it scales with sentence length. In particular, the label set size for projective trees is $\Theta(n^2)$, as a label can have any combination of $\Theta(n)$ {\tcl} symbols and $\Theta(n)$ {\tor} symbols adding up to at most $n-1$ total such symbols, denoting outgoing left and right arcs. In contrast, newer bounded encodings like hexatagging~\citep{amini-etal-2023-hexatagging} use a constant number of labels. To address this, \citet{gomez-rodriguez-etal-2023-4} proposed the 4-bit encoding, a compact bracketing scheme for projective trees with 16 fixed labels.

We next define a framework to compress bracketing encodings, based on \citet{yli-jyra-2019-transition}'s notion of \emph{rope cover}, of which we will show that the 4-bit encoding is a particular case. However, the 4-bit encoding is not optimal in bracket usage, so we propose an encoding that represents any projective tree with 12 labels instead of 16.

\subsection{Hierarchical bracketing framework}\label{sec:hierarchical}

To avoid unbounded strings of brackets in dependency labels, we establish a hierarchy of two levels of dependency arcs, yielding a hierarchical bracketing scheme with two levels of brackets: superbrackets and semibrackets~\citep{yli-jyra-2019-transition,superjlm}.

We first illustrate this by example. Consider the dependency tree in Figure~\ref{subfig:naive}. We observe that the leftmost part (nodes from 0 to 4) is composed of an arc  $0 \rightarrow 4$ and  a number of arcs that go inwards from its two endpoints. Taking advantage of this, we declare $0 \rightarrow 4$ to be a \emph{structural arc}, and we encode it with two \emph{superbrackets} {\sor} and {\scr} (with the complete set of superbrackets being $\{ {\sol}, {\sor}, {\scl}, {\scr} \}$). 

All the other arcs between nodes 0 and 4 \emph{lean on} the structural arc $0 \to 4$, in the sense that they are covered by it and they share one of its endpoints. We call them \emph{auxiliary arcs} associated with that structural arc. Exploiting this, we encode each of these arcs with a single symbol, called a \emph{semibracket}, which we place in the label of the endpoint that is not shared with the structural arc. For example, to encode the left arc $3 \leftarrow 4$, we just place a left semibracket {\bol} in the label of node 3, indicating that it has an incoming auxiliary arc from the right. A bracket at the other endpoint (4) is not needed, because the arc, being auxiliary, is supposed to lean on a structural arc. Thus, the right endpoint is understood to be at the nearest closing superbracket to the right (note that it does not matter whether this superbracket is of type {\scl} or {\scr}). Analogously, the arc $0 \rightarrow$ 1 only needs a semibracket {\bcr} with the left endpoint being the closest opening superbracket to the left, in this case {\sor}. 

This leads to the encoding in Figure~\ref{subfig:optimal}. The nodes of the second part of the tree (4 to 7) are assigned with super and semibrackets following a similar procedure. The auxiliary arc $6 \leftarrow 7$ leans on the structural arc $4 \to 7$, just like $3 \leftarrow 4$ did on $0\to 4$, so $4\to 7$ is also encoded with {\sor} and {\scr} while $6 \leftarrow 7$ only needs {\bol} in node $6$, with the right endpoint being given by
{\scr} in node 7. The arc $5\leftarrow 6$ does not lean on any other arc of the tree, so we mark it as another structural arc.

As observed, this means that we can encode the tree with fewer brackets than the original bracketing encoding (10 vs. 14, cf. Figures~\ref{subfig:naive} and \ref{subfig:optimal} counting the number of individual bracket symbols). The amount of brackets saved depends on which arcs we mark as structural, as each structural arc needs two brackets while auxiliary arcs are represented with just one. For example, Figure~\ref{subfig:hierarchical} shows an alternative where two more arcs are marked as structural, leading to using two more brackets. Thus, minimizing the number of brackets needed requires finding the minimum possible set of structural arcs such that all other arcs lean on them. This yields the \emph{optimal hierarchical bracketing}, which in the example is the one in the center.

\paragraph{Formalization}

We now define the framework, following the definitions of~\citet{yli-jyra-2019-transition}.\footnote{Due to differences in scope, we do not literally follow the terminology of~\citet{yli-jyra-2019-transition}, but make a substantive adaptation for our purposes. We skip definitions that we do not need, add some extra terms (most importantly structural and auxiliary arcs, which are present in~\citep{yli-jyra-2019-transition} but were not given an explicit name), change the perspective on others, simplify the bracketing notation, and consider non-optimal bracketings which were not contemplated by~\citeauthor{yli-jyra-2019-transition}.} We say that an arc $(i,j)$ \emph{supports} $(k,l)$ (or equivalently, $(k,l)$ \emph{leans} on $(i,j)$) if $(i,j)$ covers $(k,l)$ and either $\min(i,j) = \min(k,l)$ or $\max(i,j) = \max(k,l)$. For example, in the tree in Figure~\ref{fig:simpletree}, $0 \rightarrow 4$ supports $0 \rightarrow 1$, $2 \leftarrow 4$ and $3 \leftarrow 4$.

A subset $R \subseteq E$ is a \emph{rope cover} of a dependency graph $G=(V,E)$ if every arc in $E \setminus R$ leans on at least one arc in $R$. Given such a rope cover, we call the arcs in $R$ \emph{structural arcs}, and the arcs in $E \setminus R$ \emph{auxiliary arcs}. $R$ is said to be a \emph{proper rope cover} if no arc in $R$ leans on another arc of $R$. ~\citet{yli-jyra-2019-transition} proves that the proper rope cover of a graph is unique.\footnote{There is a specific case where it is not unique: graphs with a cycle of length 2, i.e. arcs $a \rightarrow b$ and $a \leftarrow b$, where either could be structural yielding two proper rope covers. This case is overlooked in~\citep{yli-jyra-2019-transition} as the proof is based on the underlying undirected graph of the dependency graph. However, this cannot happen in trees, and even in graphs, it can easily be solved by setting an arbitrary criterion in the definition (i.e., to force the right arc to be the structural one in this situation). Thus, we ignore this exception from now on.} In Figures~\ref{subfig:optimal}  and \ref{subfig:hierarchical}, the subsets of red, solid arcs are rope covers of the example tree, with the one in~\ref{subfig:optimal} being the proper rope cover.

\paragraph{Encoding} Given a dependency graph $G=(V,E)$ where no arcs cross, and a rope cover $R \subseteq E$, we can encode the graph as follows (and we call this encoding the \emph{hierarchical bracketing encoding of $G$ induced by $R$}): for each rightward arc $w_i \rightarrow w_j \in R$, we add an opening superbracket {\sor} to the label of $w_i$ and a closing superbracket {\scr} to the label of $w_j$. For each leftward arc $w_i \leftarrow w_j \in R$, we add an opening superbracket {\sol} to the label of $w_i$ and a closing superbracket {\scl} to the label of $w_j$. For an arc $(w_i,w_j) \notin R$, if it leans on a structural arc with left (right) endpoint $w_i$, we add to the label of $w_j$ the semibracket {\bcr} ({\bol}). Otherwise, it will lean on a structural arc with left (right) endpoint $w_j$, and we add to the label of $w_i$ the semibracket {\bcl} ({\bor}). Regarding the ordering of brackets within the same label, we always arrange them in such a way that closing brackets appear first (in increasing order of arc length) followed by opening brackets (in decreasing order of arc length), to ensure correct nesting.

\paragraph{Decoding} To decode, we use Algorithm~\ref{algo:decoding}. The algorithm uses a stack $S$ to process brackets and put the resulting arcs into the arc set $A$. Labels and symbols therein are read from left to right. When an opening bracket (of any kind) is found, it is pushed to the stack, keeping track of the index of the label where it appeared. When a closing semibracket is read, it is matched to the top node on the stack (which can be any opening superbracket), creating the corresponding arc, but without removing said superbracket from the stack. Finally, when a closing superbracket is found, it is matched to any semibrackets on top of the stack until finding the first superbracket, which is also matched. All these matched nodes are removed from the stack, and the corresponding arcs created.

\begin{algorithm}
\caption{Decoding for Noncrossing Graphs}
\begin{small}
\begin{algorithmic}[1]
\Procedure{Decode}{$l_1, \dots, l_n$}
    \State $S \gets$ empty stack ; $A \gets$ empty set
    \For{$i = 1 \to n$}
        \For{symbol in $l_i$}
            \If{symbol $\in \{{\bol}, {\bor}, {\sol}, {\sor}\}$}
                \State \textsc{Push}($S, (\text{symbol}, i)$)
            \ElsIf{symbol $= {\bcr}$}
                \State $(s, j) \gets \textsc{Peek}(S)$ \Comment{$s \in \{ {\sor},{\sol} \}$}
                \State $A \gets A \cup \{ j \rightarrow i \}$
            \ElsIf{symbol $= {\bcl}$}
                \State $(s, j) \gets \textsc{Peek}(S)$ \Comment{$s \in \{ {\sor},{\sol} \}$}
                \State $A \gets A \cup \{ j \leftarrow i \}$
            \ElsIf{symbol $={\scr}$ \textbf{or} symbol $={\scl}$}
                \State \Call{CloseSuperbracket}{symbol, i, S, A}
            \EndIf
        \EndFor
    \EndFor
\EndProcedure
\newline
\Procedure{CloseSuperbracket}{rbracket, i, S, A}
    \State $(\text{lbracket}, j) \gets \textsc{Pop}(S)$
    \While{ lbracket $={\bol}$ \textbf{or} lbracket $={\bor}$ }
        \If{lbracket $={\bol}$} $A \gets A \cup \{ j \leftarrow i \}$
        \Else \  $A \gets A \cup \{ j \rightarrow i \}$ \Comment{lbracket $={\bor}$}
        \EndIf
        \State $(\text{lbracket}, j) \gets \textsc{Pop}(S)$
    \EndWhile
    \If{lbracket $={\sol}$ \textbf{and} rbracket $={\scl}$}
        \State $A \gets A \cup \{ j \leftarrow i \}$
    \ElsIf{lbracket $={\sor}$ \textbf{and} rbracket $={\scr}$}
        \State $A \gets A \cup \{ j \rightarrow i \}$
    \EndIf
\EndProcedure
\end{algorithmic}
\end{small}
\label{algo:decoding}
\end{algorithm}

The complexity of Algorithm~\ref{algo:decoding} is linear with respect to the number of arcs in the graph. This can be shown by observing that, for each arc, we perform exactly one push to the stack (when the left bracket is read); as well as exactly one pop or one peek, depending on whether the right bracket is a superbracket or a semibracket. Thus, when operating on dependency trees, complexity with respect to sentence length is $O(n)$, as there is one arc per word; whereas for dense graphs (not often found in NLP) complexity can increase to $O(n^2)$.

\paragraph{Ensuring well-formedness} Like all parsing-as-sequence-labeling approaches, hierarchical bracketing encodings are not surjective, so practical decoding implementations need to be able to deal with ill-formed label sequences (for example, those with mismatched brackets). In our implementation of Algorithm~\ref{algo:decoding}, we follow common practice in sequence labeling parsing, by which illegal label sequences are handled by very simple heuristics. In particular, if brackets fail to match, we just match matching brackets and ignore any extra brackets -- unclosed opening brackets remain in the stack after Algorithm~\ref{algo:decoding} and not create any arcs, and unmatched closing brackets (e.g. when we try to pop the stack in line 16 or 20, but it's empty) are discarded. If the desired output is a tree, cycles and reentrancies are avoided by checking for them whenever the algorithm adds a new arc to $A$ (which can be done in inverse Ackermann time using path compression and union by rank, i.e., it does not affect the computational complexity of the decoding algorithm) and nodes without a parent are attached to the syntactic root in post-processing.

\paragraph{Compact hierarchical bracketing} The hierarchical bracketing framework we have just introduced can be used to define different encodings, depending on which criteria we use to set the structural arcs (i.e., to obtain a rope cover). In the worst case, the resulting encodings can still be unbounded: for example, if we consider the degenerate case of a rope cover where $R = E$ (i.e., all arcs are structural), the resulting encoding is equivalent to the naive bracketing of~\citep{strzyz-etal-2019-viable,strzyz-etal-2020-bracketing}.

We now define a sufficient condition to obtain bounded encodings. Given a dependency graph $G=(V,E)$, and a rope cover $R \subseteq E$, we say that $R$ is a \emph{compact rope cover} if no node in $V$ is a head of more than one structural arc going to the same direction, and no node is a dependent of more than one structural arc coming from the same direction. This condition limits the amount of superbrackets of each kind that can appear in the same label to 1 (e.g., we cannot have two {\sor} on a label because we cannot have more than one outgoing structural arc to the right). Since the amount of semibrackets of each kind that can appear in the same label is always at most 1 regardless of the rope cover used (semibrackets always match the closest superbracket and do not modify the stack, so a group of equal semibrackets would all create the same arc); a compact rope cover will induce an encoding where the length of each label is bounded by a constant, which is thus bounded. We call the encoding induced by a compact rope cover a \emph{compact hierarchical bracketing encoding}.

\subsection{Properties of hierarchical brackets for projective trees}

The framework described in Section~\ref{sec:hierarchical} works for any dependency graph without crossing arcs. However, it is especially interesting to focus on projective dependency trees, as their properties guarantee a particularly compact label set. In particular, we can show the following result:

\newtheorem{theorem}{Theorem}

\begin{theorem}
Any compact hierarchical bracketing encoding for projective dependency trees uses at most 16 labels.
\label{theo:sixteenlabel}
\end{theorem}

To prove it, we consider the following:
\begin{itemize}
\item Since each node in a dependency tree has exactly one parent, each label for a dependency tree (projective or not) must have exactly one bracket from the set $\{ {\bol}, {\sol}, {\bcr}, {\scr} \}$.
\item The encoding for a projective dependency tree cannot have any brackets of the form {\bcl} or {\bor}. {\bcl} can only appear if a leftward arc leans on an arc that covers it. In a tree, the covering arc must necessarily be a rightward arc (so the single-head constraint is not violated) and this means that the node with the bracket {\bcl} dominates over the covering arc, which cannot happen in a projective tree. The same reasoning (but mirrored) applies to {\bor}.
\item We already established in the general framework that closing brackets appear before opening brackets in labels. In projective trees, we can further observe that leftward closing brackets ({\scl}) must appear before rightward ones ({\bcr} or {\scr}), and leftward opening brackets ({\bol} or {\sol}) before rightward ones ({\sor}). This is for similar reasons as the previous point: for example, a label with {\bcr} or {\scr} before {\scl} would mean that the corresponding node has an incoming arc from the left and a \emph{longer} outgoing arc to the right. In the context of a tree, this means that the node dominates an arc that covers it, which is forbidden in projective trees. Analogous reasoning can be done for the other forbidden combinations.
\end{itemize}

Taking all these points into account, we conclude that the labels of a compact hierarchical bracketing encoding for projective dependency trees must necessarily be of the form ({\scl})? ({\bcr} | {\scr} | {\bol} | {\sol}) ({\sor})?

Counting the labels allowed by this expression, we directly derive Theorem~\ref{theo:sixteenlabel}.

\subsection{The 4-bit encoding as hierarchical bracketing}

\citet{gomez-rodriguez-etal-2023-4} present the \emph{4-bit encoding}: a 16-label encoding for a superset of projective dependency trees, where each label is composed of 4 bits ($b_0b_1b_2b_3$): $b_0$ indicates whether the corresponding node is a left or right dependent, $b_1$ whether it is the farthest left or right dependent, and $b_2$ ($b_3$) whether it has any left (right) dependents.

This encoding can be seen as a compact hierarchical bracketing for projective trees. To see it, for a given projective tree $T=(V,E)$, consider the rope cover $R_{4b}$ built by taking the longest rightward arc and longest leftward arc going out of each node. Under this rope cover, the bracket from the set $\{ {\bol}, {\sol}, {\bcr}, {\scr} \}$ associated to a given word will be a superbracket if, and only if, that word is the farthest left or right dependent of its head (i.e., if $b_1 = 1$). Since $b_0$ determines direction, $b_0$ and $b_1$ together select exactly one bracket from that set. In turn, presence of ${\scl}$ in a label corresponds to $b_2=1$ (indicating having left dependents, which means we must use ${\scl}$ for the origin of the corresponding left arc(s) since ${\bcl}$ cannot appear in projective trees, as argued earlier) and the same applies to ${\sor}$ and $b_3=1$. Thus, 4-bit labels are isomorphic to the labels of the compact hierarchical bracketing encoding induced by the rope cover $R_{4b}$, with the correspondence shown in Table~\ref{tab:4bit}. A simple example of a tree with its rope cover $R_{4b}$ and the induced labeling, corresponding to the 4-bit encoding, is shown in Figure~\ref{subfig:hierarchical}.

\begin{table}[tbp]
\centering
\begin{tabular}{c|cc|cc|}
\hline
\backslashbox{$b_0b_1$}{$b_2b_3$} & 00 & 01 & 10 & 11 \\ \hline
00 & {\bol} & {\bol} {\sor} & {\scl} {\bol} & {\scl} {\bol} {\sor} \\ \hline
01 & {\sol} & {\sol} {\sor} & {\scl} {\sol} & {\scl} {\sol} {\sor} \\ \hline
10 & {\bcr} & {\bcr} {\sor} & {\scl} {\bcr} & {\scl} {\bcr} {\sor} \\ \hline
11 & {\scr} & {\scr} {\sor} & {\scl} {\scr} & {\scl} {\scr} {\sor} \\ \hline
\end{tabular}
\caption{Conversion of the 4-bit labels \cite{gomez-rodriguez-etal-2023-4} to hierarchical bracketing labels.}
\label{tab:4bit}
\end{table}

\subsection{Optimal hierarchical bracketing}

By viewing the 4-bit encoding through the lens of the hierarchical bracketing framework, we can notice that it is suboptimal in terms of number of brackets needed to encode a tree. An example can be seen in Figure~\ref{fig:simpletree}, where the 4-bit encoding yields the rope cover and labels shown on the right; but the alternative in the center has a rope cover with two fewer structural arcs (and thus, uses two fewer brackets) to encode the same tree. This begs the question of whether it is possible to define an \emph{optimal hierarchical bracketing} for projective trees, i.e. one that guarantees encoding each tree with the minimum possible amount of brackets within hierarchical bracketing encodings.

This can be done by computing the proper rope cover. \citet{yli-jyra-2019-transition} provides a simple algorithm for this purpose, which works for any dependency graph $G=(V,E)$: 
\begin{enumerate}
\item All arcs in $E$ start unmarked. Among the unmarked arcs with the leftmost left endpoint, take the longest one and mark it as structural arc, thus adding it to the proper rope cover.
\item Mark all the arcs that lean on it as auxiliary.
\item Repeat the process until every arc is marked.
\end{enumerate}

We will now show the following previously unproven property of the proper rope cover:

\begin{theorem}
The proper rope cover of a dependency graph $G=(V,E)$ has minimum cardinality among rope covers of $G$.
\label{theo:prcminimum}
\end{theorem}

To prove this, we can proceed as follows. Let $R_{pr}$ be the unique proper rope cover of $G$. We claim that $R_{pr}$ must be of minimum cardinality among all rope covers. Equivalently, we can show that \emph{any} rope cover $R \subseteq E$ can be transformed into some rope cover $R'$ (possibly itself) that is proper with $|R'| \le |R|$.  Since the proper rope cover is unique, it follows $R' = R_{pr}$, forcing $|R_{pr}| \le |R|$. Hence $|R_{pr}|$ is minimal.

The transformation process is as follows. If $R$ is initially proper, the transformation is done. If it is not proper, then by definition there exist arcs $r_1, r_2 \in R$ such that $r_1$ leans on $r_2$. Choose such a pair of arcs. By definition, $r_1$ and $r_2$ share an endpoint $e$, and $r_2$ covers $r_1$. Without loss of generality, suppose that $e$ is the right endpoint of $r_1$ and $r_2$. We call $e_l$ the left endpoint of $r_1$.

\begin{figure}[tbp]
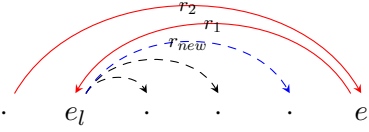

\centering
\begin{dependency}[arc edge, text only label]
   \begin{deptext}[column sep=0.24cm,minimum width=0.7cm]
      $\cdot$ \& $e_l$  \& $\cdot$ \& $\cdot$ \& $\cdot$ \& $e$ \\
   \end{deptext}
   \depedge[color=red]{1}{6}{$r_2$}
   \depedge[dashed]{2}{3}{}
   \depedge[dashed]{2}{4}{}
   \depedge[color=blue,dashed]{2}{5}{$r_{new}$}
   \depedge[color=red]{6}{2}{$r_1$}
\end{dependency}
\caption{An iteration of the process to transform a rope cover to the proper rope cover. The arc $r_1$ in the rope cover would be replaced with $r_{new}$.}
\label{fig:iteration}
\end{figure}

We will now remove $r_1$ from the rope cover, possibly replacing it with another arc. To do this, we need to take into account that all arcs in $E \setminus R$ that lean on $r_1$ need to still be supported by a structural arc in the new rope cover. Arcs can be in this situation either because their right endpoint is $e$, or their left endpoint is $e_l$. The former are supported by $r_2$, and will continue to be even if we remove $r_1$. So we only need to ensure that any arcs with left endpoint $e_1$ (and covered by $r_1$) have support. For this purpose, if there are arcs in this situation, we take the longest and add it to the rope cover to replace $r_1$ (if it was not already in the rope cover). This arc will cover the remaining arcs, so the obtained arc set is a new rope cover. This process is illustrated graphically in Figure~\ref{fig:iteration}.

This process either preserves the size of the rope cover (if $r_1$ is replaced with another arc) or decreases it by $1$. Also note that it strictly reduces the sum of edge lengths of the rope cover (even if no arc is removed, $r_1$ is replaced with a strictly shorter arc covered by it). Thus, we can iterate it while there is a pair of leaning arcs, with the assurance that it will finish, and the result will be the proper rope cover. This concludes the proof of Theorem~\ref{theo:prcminimum}.

From this, we can derive the following result:

\newtheorem{corollary}{Corollary}

\begin{corollary}
The hierarchical bracketing induced by the proper rope cover is optimal.
\label{coro:prcminimum}
\end{corollary}

This is shown by observing that, in the hierarchical bracketing encoding induced by a rope cover, each structural arc generates two superbrackets, whereas each auxiliary arc generates one semibracket. By Theorem~\ref{theo:prcminimum}, the proper rope cover has the minimum possible amount of structural arcs, thus, it also minimizes the total number of brackets generated for a given graph.

Hence, with the above algorithm by~\citet{yli-jyra-2019-transition} to compute the proper rope cover, we get a hierarchical bracketing which is optimal in terms of number of brackets used to encode each given tree. We will now prove that, when applied to projective trees, it also reduces the number of labels with respect to alternatives like e.g. the 4-bit encoding:

\begin{theorem}
When applied to projective trees, the optimal hierarchical bracketing encoding induced by the proper rope cover uses at most 12 labels.
\label{theo:twelvelabel}
\end{theorem}

\begin{figure*}[btp]
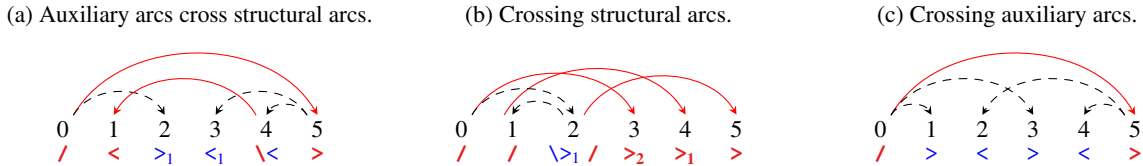
\small\centering 
    \begin{subfigure}[t]{0.3\textwidth}\centering
        \caption{\label{subfig:nonprojsemi}Auxiliary arcs cross structural arcs.}
        \begin{dependency}[arc edge,hide label]
           \begin{deptext}[column sep=0.12cm,minimum width=0.55cm]
              0 \& 1 \& 2 \& 3 \& 4 \& 5 \\
              \sor \& \sol \& \bcri{1} \& \boli{1} \& \scl \bol \& \scr \\
           \end{deptext}
           \depedge[color=red]{1}{6}{}
           \depedge[dashed]{1}{3}{}
           \depedge[dashed]{6}{4}{}
           \depedge[dashed]{6}{5}{}
           \depedge[color=red]{5}{2}{}
        \end{dependency}
    \end{subfigure}\hspace{0.5cm}
    \begin{subfigure}[t]{0.3\textwidth}\centering 
        \caption{\label{subfig:nonprojsuper}Crossing structural arcs.}
        \vspace{0.87em}
        \begin{dependency}[arc edge,hide label]
           \begin{deptext}[column sep=0.12cm,minimum width=0.55cm]
              0 \& 1 \& 2 \& 3 \& 4 \& 5 \\
              \sor \& \sor \& \bcl \bcri{1} \sor \& \scri{2} \& \scri{1} \& \scr \\
           \end{deptext}
           \depedge[color=red]{1}{4}{}
           \depedge[color=red, edge horizontal padding={-3pt}]{2}{5}{}
           \depedge[color=red]{3}{6}{}
           \depedge[dashed]{1}{3}{}
           \depedge[dashed]{3}{2}{}
        \end{dependency}
    \end{subfigure}\hspace{0.5cm}
    \begin{subfigure}[t]{0.3\textwidth}\centering
        \caption{\label{subfig:fakeproj}Crossing auxiliary arcs.}
        \begin{dependency}[arc edge,hide label]
           \begin{deptext}[column sep=0.12cm,minimum width=0.55cm]
              0 \& 1 \& 2 \& 3 \& 4 \& 5 \\
              \sor \& \bcr \& \bol \& \bcr \& \bol \& \scr \\
           \end{deptext}
           \depedge[color=red]{1}{6}{}
           \depedge[dashed]{1}{2}{}
           \depedge[dashed]{1}{4}{}
           \depedge[dashed]{6}{5}{}
           \depedge[dashed]{6}{3}{}
        \end{dependency}
    \end{subfigure}
    \caption{\label{fig:nonproj} Examples of the non-projective extension. In Figure \ref{subfig:nonprojsemi},
    we add indexes to semibrackets to skip the matching superbracket. In Figure \ref{subfig:nonprojsuper},     indexes are also added to superbrackets to skip multiple opening superbrackets. In Figure \ref{subfig:fakeproj}, no indexes are needed when auxiliary arcs cross other auxiliary arcs.    }
\end{figure*}

To prove this, consider the 16 possible labels of compact hierarchical bracketings (Theorem~\ref{theo:sixteenlabel}), which are also shown explicitly in Table~\ref{tab:4bit} (since they are also the labels for the 4-bit encoding). Out of those 16 labels, there are four that cannot appear in the bracketing induced by the proper rope cover: ${\sol} {\sor}$, ${\scl} {\scr}$, ${\scl} {\sol} {\sor}$ and ${\scl} {\scr} {\sor}$. This is because presence of a label including ${\sol} {\sor}$ indicates that its corresponding node has both an incoming structural arc from the right and an outgoing structural arc to the right and, by construction, one would lean on the other, which is not allowed in the proper rope cover. The same reasoning can be made for the combination ${\scl} {\scr}$.

Thus, we have derived a hierarchical bracketing encoding for projective trees that is easily obtainable and reduces the label set from $16$ to $12$ when compared with the 4-bit encoding, while sharing common underlying principles.

\section{Non-projectivity}

Our above definitions of hierarchical bracketing are for dependency graphs without crossing arcs. We now add the possibility of crossing arcs (and thus support non-projective trees). One option to do so would be to use multiplanarity (dividing the graph into subsets without crossings) as in~\citep{strzyz-etal-2020-bracketing,gomez-rodriguez-etal-2023-4}. We choose not to follow this path here as it is not a straightforward match with our single-stack decoding process, and instead, again inspired by~\citet{yli-jyra-2019-transition}, we use brackets that can skip others when matching.

We make the next extensions to the framework:
\begin{enumerate}
    \item Without crossing arcs, by construction, a semibracket always represents an arc whose other endpoint leans on the shortest structural arc covering the semibracket node. With crossing arcs, the other endpoint can be on a more external structural arc. If a semibracket is covered by several structural arcs and it encodes an arc leaning on the $i$th covering structural arc, we add an index $i-1$ to it (meaning that it will skip $i-1$ superbrackets when matching). An example can be seen in Figure~\ref{subfig:nonprojsemi}.
    \item Structural arcs can cross other structural arcs. For this, we add an index $i$ to a right superbracket ({\scr},{\scl}) to indicate when it needs to skip $i$ superbrackets on the stack (i.e., match the $(i+1)$th matching superbracket on the stack instead of the first), as in 
    Figure~\ref{subfig:nonprojsuper}.
\end{enumerate}

\begin{figure}[btp]\centering\small 
    \begin{dependency}[text only label, arc edge, label style={above}]
        \begin{deptext}[column sep=10pt]
            1 \& 2 \& 3 \& 4 \& 5 \& 6 \& 7 \& 8 \& 9 \\[0.7em] 
            \sor \& \bol \& \bcr \& \bol \&  \&  \& \scr \& \& \& \\[0.7em] 
            \sor \& \bol \& \bcr\sor \& \bol \& \boli{1} \&  \& \scri{1} \& \scr \& \\[0.7em] 
            \sor \& \bol \& \bcr\sor \& \sor\bol \& \boli{1} \& \boli{2} \& \scri{2} \& \scri{1} \& \scr \\ 
        \end{deptext}
        % structural arcs 
        \depedge[edge style={red}]{1}{7}{}
        \depedge[edge style={red}]{3}{8}{}
        \depedge[edge style={red}, edge horizontal padding=-3pt]{4}{9}{}
        % auxiliar arcs 
        \depedge[dashed]{1}{3}{}
        \depedge[dashed]{7}{2}{}
        \depedge[dashed]{7}{4}{}
        \depedge[dashed]{8}{5}{}
        \depedge[dashed]{9}{6}{}
    \end{dependency}
    \caption{\label{fig:encoding-steps}Step-by-step encoding of a non-projective tree. The bracketing sequence of the first row encodes the structural arc $(1\to 7)$ with its auxiliary arcs $(1\to 3)$, $(7\to 2)$ and $(7\to 4)$ The second row encodes $(3\to 8)$ with $(8\to 5)$, increasing the index of the $5$th and $7$th bracket to avoid decoding $(7\to 5)$ or $(3\to 7)$. The third row encodes $(4\to 9)$ with $(9\to 6)$, and increases twice the index of the 6th bracket, since  $(6\to 9)$ crosses with two structural arcs $(1\to 7)$ and $(3\to 8)$; and the 7th and 8th brackets, since they both cross with $(6\to 9)$.}
\end{figure}

These extensions are sufficient to support every non-projective graph -- note that auxiliary arcs can also cross other auxiliary arcs, as in Figure~\ref{subfig:fakeproj}, but no extension to labels is needed to support this: it is enough if the decoding algorithm ignores any intervening opening semibrackets on the stack when matching a closing semibracket (i.e., modifying lines 8 and 11 of Algorithm~\ref{algo:decoding} to fetch the first superbracket instead of peeking).

Encoding is done step by step, by incrementally adding to the labels a structural arc and its associated auxiliary arcs at each step. Figure \ref{fig:encoding-steps} shows a step-by-step example of this process.

To support the extension in the decoding algorithm, apart from the above change, it suffices to implement the following changes to Algorithm~\ref{algo:decoding}: (1) for closing semibrackets indexed with $i$, lines 8 and 11 fetch the $(i+1)$th opening superbracket from the stack instead of the first, (2) for closing superbrackets indexed with $i$, procedure CloseSuperbracket matches the $(i+1)$th opening superbracket instead of the first (behavior with intervening opening semibrackets is the same), and (3) for opening semibrackets indexed with $i$, when matched (line 17) no arcs are created, instead they are placed back on the stack with their index decreased by one (and removed if it becomes zero). The resulting pseudocode is in Appendix \ref{ap:non-projective-decoding-example} (Algorithm~\ref{algo:decodingarbitrary}), along with an example of an encoded sentence (Figure~\ref{fig:complex-nonprojective-tree}).

The bracket indexes are bounded by the maximum number of structural arcs that cover a fencepost between two nodes, minus one. For the optimal hierarchical bracketing, that maximum number (before subtracting one) is called \emph{rope thickness}~\citep{yli-jyra-2019-transition}, and coincides with the maximum number of open superbrackets that appear in the stack at the same time. \citet{yli-jyra-2019-transition} shows that trees in UD 2.4 have rope thickness at most 8. This does not mean that we actually need brackets with index 7 to represent all trees in UD: rope thickness gives a conservative upper bound, as most of the time having $k$ brackets in the stack does not mean that we need to skip to the deepest to encode the tree. In the set of treebanks used in our experiments, index 2 gives full coverage for every language except Ancient Greek.

\begin{table}[tbp]\centering\scriptsize
    \setlength{\tabcolsep}{1.6pt}
    \begin{tabular}{|c|c|ccccc|cc|cccc|}
        \cline{2-13}
        \multicolumn{1}{c|}{} & \multirow{2}{*}{\textbf{\#trees}} & \multicolumn{5}{c|}{\textbf{\#labels}} & \multicolumn{2}{c|}{\textbf{\#rels}} &\multicolumn{4}{c|}{\textbf{\#indices}}\\ 
        \multicolumn{1}{c|}{} & & \textbf{B}\SB{4}& \textbf{O}\SB{P} & \textbf{B}\SB{7} & \textbf{O}\SB{NP} & \textbf{H}\textsuperscript{+}& \textbf{O}\SB{P} & \textbf{O}\SPSB{+}{P}  & \textbf{0} & \textbf{1} & \textbf{2} &  \textbf{\boldmath$\geq 3$} \\ 
         \hline 
        \textit{grc} & 13.9k & 16 & 12 & 103 & 168 & 8 &26 & 248 & 42.87 & 51.32 & 5.50 & 0.25\\
        \textit{en} & 16.6  & 16 & 12 & 60 & 55 & 8 & 53 & 169 &  97.98 & 1.98 & 0.04 &0\\
        \textit{fi} & 15.1k & 16 & 12 & 59 & 56 & 8 & 47 & 150 &  96.66 & 2.40 & 0.93 &0\\
        \textit{fr} & 16.3k  & 16 & 12 & 51 & 50 & 8 & 56 & 142 &  96.63 & 3.23 & 0.14 &0\\
        \textit{he} & 6.1k & 16 & 12 & 37 & 30  & 8& 37 & 52 &  99.22 & 0.75 & 0.03 & 0\\
        \textit{ru} & 5k & 16 & 12 & 60 & 52  &8& 42& 153 & 95.27 & 4.31 & 0.42 & 0\\
        \textit{ta} & 600 & 16 & 12 & 24 & 17& 8 & 29  & 35 & 98.33 & 1.67 & 0 & 0\\
        \textit{ug} & 3.5k & 16 & 12 & 40 & 35  & 8 & 40& 67  & 93.40 & 6.57 & 0.03 & 0\\
        \textit{wo} & 2.1k & 16 & 12 & 40 & 27 &8  & 38&  62 & 97.15 & 2.85 & 0 & 0\\
        \hline 
        PTB & 44k & 16 & 12 & 22 & 21 & 8 & 45 & 46 & 99.90 & 0.10 & 0 & 0\\
        \hline 
    \end{tabular}
    \caption{\label{tab:stats}Treebank and encoding statistics. Number of trees in the dataset (\textbf{\#trees}), generated labels (\textbf{\#labels}) and dependency types (\textbf{\#rels}), and 
    \% of trees that require indices $p\in\{0,1,2,\geq3\}$ in non-projective labels. The notation is the following: \textbf{B}\textsubscript{4} and \textbf{B}\textsubscript{7} for the 4 and 7-bit encoding \cite{gomez-rodriguez-etal-2023-4}, respectively; \textbf{O}\textsubscript{P} (and \textbf{O}\textsubscript{NP}) for the projective (and non-projective) optimal hierarchical bracketing; \textbf{H} for the hexatagging parser \cite{amini-etal-2023-hexatagging}; and the symbol ($^+$) denotes if pseudo-projectivity is used.}
\end{table}

\section{Experiments}
We conducted experiments in the Penn Treebank \cite[][PTB]{marcus-etal-1993-building} and 9 different languages from UD 2.14 \cite{nivre-etal-2020-universal} to analyze the performance of our encodings in a diverse set of tree structures: Ancient Greek, English, Finnish, French, Hebrew, Russian, Tamil, Uyghur and Wolof. We report in this section the LAS and LCM  metrics \cite{nivre-etal-2007-conll} and in  Appendix \ref{subsec:ap-results} the detailed results. All our code is available at \href{https://github.com/anaezquerro/separ}{\texttt{github.com/anaezquerro/separ}}.

\begin{table*}[tbp]\centering\small
    \setlength{\tabcolsep}{3pt}
        \begin{tabular}{|c|cc|cc|cc|cc||cc|cc||cc|cc|}
        \cline{2-17}
        \multicolumn{1}{c}{}& \multicolumn{2}{|c|}{\textbf{B}\textsubscript{4}} & \multicolumn{2}{c|}{\textbf{O}\textsubscript{P}} & \multicolumn{2}{c|}{\textbf{B}\SPSB{+}{4}} &\multicolumn{2}{c||}{\textbf{O}\SPSB{+}{P}}& \multicolumn{2}{c|}{\textbf{B}\textsubscript{7}} & \multicolumn{2}{c||}{\textbf{O}\textsubscript{NP}} &  \multicolumn{2}{c|}{\textbf{H}\textsuperscript{+}}& \multicolumn{2}{c|}{\textbf{DM}}  \\
        \hline 
        \textit{grc} & 59.64 & 8.81 & 57.70 & 8.96 & 63.02 & 9.95 & \underline{63.82} & \underline{11.10} & \underline{62.59} & 10.34 & 62.46 & \underline{11.10} & 60.74 & 9.42 & 70.79 & 13.40 \\
        \textit{en}  & 91.99 & 65.43 & 92.00 & 64.37 & \underline{92.50} & \underline{66.15} & 92.45 & 65.72 &  \underline{92.68} & \underline{65.24} & 91.69 & 65.00 & 91.75 & 63.75 & 93.31 & 67.02 \\
        \textit{fi}  & 88.04 & 42.38 & 88.13 & 44.31 & \underline{89.32} & \underline{45.98}  & 88.65 & 44.74 &  \underline{89.11} & \underline{45.27} & 88.58 & \underline{45.27} & 88.65 & 44.76 & 81.87 & 0.00 \\
        \textit{fr}  & 91.90 & 34.86 & 91.40 & 35.58 & \underline{92.19} & \underline{36.30} & 92.06 & 35.34 &  \underline{91.82} & 35.10 & 91.59 & 37.02 & 91.50 & 33.41 & 93.57 & 39.18 \\
        \textit{he}  & \underline{90.14} & \underline{30.35} & 89.51 & \underline{30.35} & 89.51 & 28.92 & 89.21 & 27.09 &  89.59 & \underline{28.92} & \underline{89.61} & \underline{28.92} & 87.86 & 25.46 & 92.02 & 33.81 \\
        \textit{ru}  & 87.99 & \underline{31.11} & 87.82 & 30.62 & 88.09 & 28.79 & \underline{88.18} & 30.95 &  \underline{88.40} & \underline{30.95} & 87.06 & 28.79 & 88.71  & 30.62 & 90.60 & 36.44 \\
        \textit{ta}  & 65.54 & 2.50 & 64.80 & \underline{5.00} & \underline{66.57} & 2.50 & 66.06 & 1.67   & \underline{68.12} & \underline{5.00} & 65.26 & 3.33 & 65.13 & 3.33 & 70.68 & 6.67 \\
        \textit{ug}  & 66.62 & 11.67 & 66.27 & 11.89 & 66.78 & 11.22 & \underline{66.88} & \underline{12.33} & \underline{66.47} & 10.33 & 65.17 & \underline{11.22} &65.56 & 11.00 & 71.39 & 14.56 \\
        \textit{wo}  & \underline{73.67} & \underline{9.15} & 70.25 & 7.66 & 73.02 & 8.30 & 71.13 & 8.30  & \underline{72.84} & \underline{11.49} & 72.37 & 8.72 & 67.23 &  5.11 & 73.56 & 10.00 \\
         \hline
         PTB & 94.76 & 51.86 &94.55 & 51.45 & \underline{94.95} & \underline{52.77} & 93.97 & 49.71 & \underline{94.61} & \underline{51.99}  & 94.45 & 51.90 & 94.80 & 51.95 & 95.33 & 51.90 \\ 
         \hline 
         \textit{avg} & 81.03 & 28.81 & 80.24 & 29.02 & \underline{81.60} & \underline{29.09} & 81.24 & 28.70 & \underline{81.62} & \underline{29.46} & 80.82 & 29.13 & 80.19 & 27.88 & 83.31 & 27.30 \\
        \hline 
    \end{tabular}
    \caption{\label{tab:results}LAS and LCM (Labeled Complete Match) performance on the test sets. Same acronyms as in Table \ref{tab:stats}, and \textbf{DM} for the biaffine parser \cite{dozat-etal-2017-biaffine}. Best projective and non-projective bracketing  encodings are underlined. Language abbreviations come from ISO 639-1.}
\end{table*}

We selected the biaffine parser \cite{dozat-etal-2017-biaffine}, and the hexatagging \cite{amini-etal-2023-hexatagging}, 4-bit and 7-bit \cite{gomez-rodriguez-etal-2023-4} encodings as baselines to assess the performance of our hierarchical encodings. For 4-bit and projective hierarchical bracketing, we also report results with the pseudo-projective transformation from \citet{nivre-nilsson-2005-pseudo}. In the case of hexatagging, we only report results with said transformation, following~\citet{amini-etal-2023-hexatagging}.\footnote{For homogeneous comparison, we use our own implementation of hexatagging, which we include as part of our code. Thus, there can be differences in hyperparameters with respect to the implementation in the original paper.}

\paragraph{Model configuration} 

We use XLM-RoBERTa \cite{conneau-etal-2020-unsupervised} as our encoder for all language treebanks, and  XLNet \cite{yang-etal-2019-xlnet} for English - both for our models and baselines - as it has shown some good results in previous work of parsing as tagging \citep{amini-etal-2023-hexatagging}. For the decoder, we rely on two feed-forward networks for label and relation prediction.\footnote{See Appendix \ref{ap:model-configuration} for more details.}

\paragraph{Results}
Table~\ref{tab:stats} shows the number of distinct labels generated by each encoding on each of treebank. For projective encodings, every treebank needs the 16 (for the 4-bit encoding) and 12 (for the optimal hierarchical bracketing) possible labels; whereas in the non-projective case the actual label usage varies per treebank. We see that the non-projective optimal hierarchical bracketing is more compact than the 7-bit encoding in every treebank except Ancient Greek (with an atypical amount of non-projectivity, since it contains poetry).

Accuracy results are in Table~\ref{tab:results}. The accuracy of our new optimal hierarchical bracketing encodings (\textbf{O}\textsubscript{P}, \textbf{O}\textsubscript{NP}) is roughly on par with the 4-bit, 7-bit and hexatagging baselines,
outperforming them in some treebanks and falling slightly behind in others. \textbf{O}\textsubscript{P/NP} seem to do better in terms of LCM than LAS (e.g., \textbf{O}\textsubscript{P} outperforms the 4-bit encoding in a majority of treebanks in LCM, but not in LAS. Similar trends, albeit less pronounced, can be observed for the non-projective encodings). This makes sense as more compactness implies less redundancy in the encoding, so while getting a complete tree correct may be easier, a mistake can affect more arcs than in a more redundant encoding. We also tried, for the first time, pseudo-projectivity~\citep{nivre-nilsson-2005-pseudo} on the 4-bit encoding and projective optimal hierarchical bracketing, which are projective encodings. The results show that it markedly boosts accuracy on non-projective treebanks, sometimes even beating native non-projective encodings.

If we compare hexatagging to all the bracketing-based encodings (including both the hierarchical encodings presented in this paper and the already-existing 4-bit and 7-bit), we can see that hexatagging excels in the PTB, but is worse on UD treebanks and especially suffers in Ancient Greek (the most non-projective treebank) and Wolof. On the other hand, the biaffine baseline obtains better LAS accuracy than every sequence-labeling encoding on most treebanks, but has the worst LCM on average.

\begin{figure}\centering
    \includegraphics[width=0.485\textwidth]{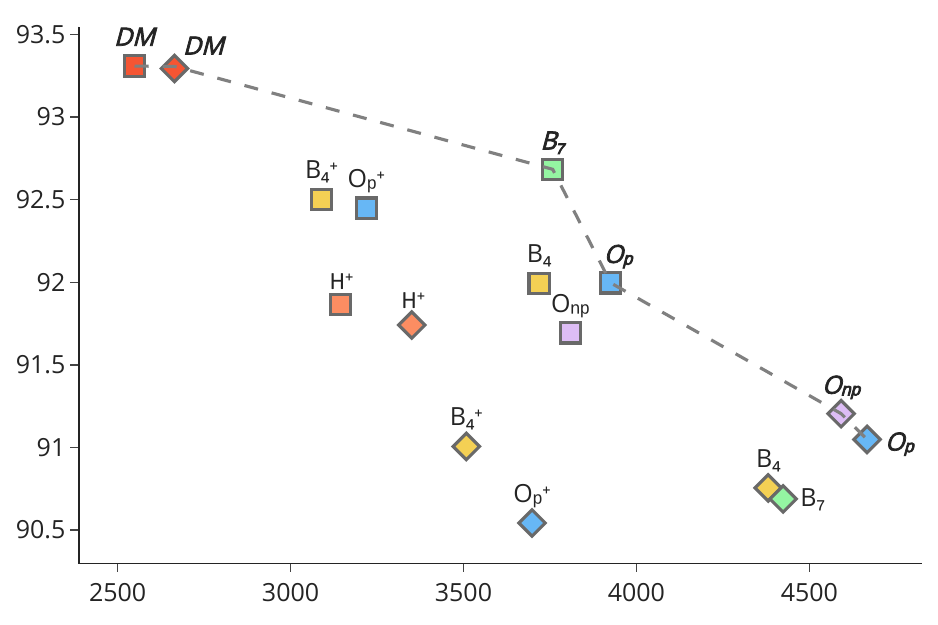}
    \caption{\label{fig:pareto}Performance (LAS, $y$-axis) vs inference speed (tokens per second, $x$-axis)  on the English-EWT test set. The Pareto front is displayed with dashed lines and highlighted in bold. Same acronyms as in Table \ref{tab:results}. Each decoder is displayed with a different color: 
    \protect\cbullet{dm}\textbf{DM}, \protect\cbullet{h}\textbf{H}\textsuperscript{+}, \protect\cbullet{b4}\textbf{B}\textsubscript{4}, \protect\cbullet{b7}\textbf{B}\textsubscript{7}, \protect\cbullet{op}\textbf{O}\textsubscript{P} and \protect\cbullet{onp}\textbf{O}\textsubscript{NP}; and each encoder with a different symbol:  XLM\protect\cdiamond{white} and XLNet\protect\csquare{white}.}
\end{figure}

Regarding efficiency, Figure~\ref{fig:pareto} shows an efficiency comparison of the encodings of Table \ref{tab:results} on the English treebank, in terms of inference speed (tokens per second) vs. accuracy (LAS), highlighting the Pareto front. As can be seen, the projective and non-projective optimal hierarchical bracketing models are noticeably faster than the 4- and 7-bit baselines, likely due to their more compact label space. While hexatagging produces even fewer labels, it needs an intermediate conversion to a binary constituency tree and this penalizes runtime with respect to optimal hierarchical bracketing, which does not require it. All sequence labeling models are faster than the biaffine parser.

\section{Conclusion}

We have advanced our understanding of bracketing encodings for dependency parsing with a framework that includes a spectrum of encodings from classic naive bracketing (all arcs are structural and require two brackets) to our novel optimal hierarchical bracketing (which minimizes the number of arcs that require two brackets), with the existing 4-bit encoding sitting in between. We have also defined and tested a new way of generalizing bracketing encodings to support non-projectivity, based on indexed brackets that skip other brackets when matching, showing that indexes bounded by 2 provide full coverage for most treebanks. Our experiments show that all our new encodings are competitive on a diverse set of treebanks. 

From a wider perspective, our encodings are a step towards more compact modeling of the full search space of dependency trees. This includes non-projective trees, which are not supported by hexatagging~\citep{amini-etal-2023-hexatagging}; and potentially general graphs, which are naturally supported by our rope cover framework. Thus, an interesting avenue for future work is to apply optimal hierarchical bracketing to parse dependency graphs.

\section*{Acknowledgments}

We acknowledge grants SCANNER-UDC (PID2020-113230RB-C21) funded by MICIU/AEI/10.13039/501100011033; GAP (PID2022-139308OA-I00) funded by MICIU/AEI/10.13039/501100011033/ and ERDF, EU; LATCHING (PID2023-147129OB-C21) funded by MICIU/AEI/10.13039/501100011033 and ERDF, EU; and TSI-100925-2023-1 funded by Ministry for Digital Transformation and Civil Service and ``NextGenerationEU'' PRTR; as well as funding by Xunta de Galicia (ED431C 2024/02), and 
CITIC, as a center accredited for excellence within the Galician University System and a member of the CIGUS Network, receives subsidies from the Department of Education, Science, Universities, and Vocational Training of the Xunta de Galicia. Additionally, it is co-financed by the EU through the FEDER Galicia 2021-27 operational program (Ref. ED431G 2023/01). We also extend our gratitude to CESGA, the supercomputing center of Galicia, for granting us access to its resources.
Furthermore, we acknowledge the Faculty of Agriculture and Forestry of the University of Helsinki, as well as projects "Theory of Computational Logics" (352420) and "XAILOG" (345612, 345633) funded by the Research Council of Finland for the continued support of the third author during the multistage writing process.

\section*{Limitations}

We adopt a default sequence labeling architecture. While the results are competitive and on par with the best-performing existing bracketing encodings, extensive fine-tuning could lead to further improvements. We refrain from doing so for two main reasons. First, our main goal is to demonstrate that the proposed encoding is effective and can be easily integrated with off-the-shelf methods without requiring extensive hyperparameter tuning. By doing so, we emphasize the practicality of our approach. Second, our computational resources are limited, consisting of six RTX 3090 GPUs, which are shared among multiple team members. Given these constraints, large-scale fine-tuning experiments would require a significant investment of time and resources, which is not feasible within our current setup. Instead, we prioritize demonstrating the effectiveness of our method under realistic conditions, where computational resources may be limited.

\section*{Ethical considerations}
Our work does not raise ethical concerns. It presents a novel approach to dependency parsing using standard treebanks, addressing primarily computational and linguistic challenges. The applications and sources involved do not include sensitive personal data, human subjects, or scenarios likely to pose ethical issues. Therefore, the findings and techniques can be adopted within the field without ethical reservations.

We do, however, acknowledge the environmental impact of training neural models, particularly in terms of CO\textsubscript{2} emissions. All experiments were conducted in Spain, and we measured the carbon footprint per epoch during both training and inference. For our models, the emission was 1.44 g CO\textsubscript{2} (training) and 0.11 g CO\textsubscript{2} (inference). These values remain relatively low compared to those of recent large-scale NLP models, reinforcing the importance of pursuing energy-efficient approaches. For context, the European Union's emissions standard for newly manufactured cars is approximately 115 g CO\textsubscript{2} per kilometer.
\bibliography{anthology,custom}

\appendix

\section{Appendix}\label{sec:appendix}

\subsection{Non-projective decoding and example}\label{ap:non-projective-decoding-example}
Algorithm \ref{algo:decodingarbitrary} shows the pseudocode of the non-projective extension for the optimal hierarchical bracketing encoding. In this case, since the modified algorithm accesses elements that are deep in the stack, the upper bound for runtime complexity is $O(|A|{i_{max}}^2)$, where $|A|$ is the number of arcs in the graph and $i_{max}$ is the maximum bracket index, as an indexed left bracket may need to be updated at most $i_{max}$ times to decrease its index while being at stack depth at most $i_{max}$. For trees, considering that setting $i_{max}=2$ has full coverage on most treebanks, this becomes $O(n)$ in practice.

In Figure \ref{fig:complex-nonprojective-tree}, we show an illustrative example of how this extension works on a graph with multiple crossing structural arcs.

\begin{figure*}[h!]\centering\small 
    \begin{dependency}[text only label, arc edge, label style={above}]
        \begin{deptext}[column sep=20pt]
            1 \& 2 \& 3 \& 4 \& 5 \& 6 \& 7 \& 8 \& 9 \& 10 \& 11 \& 12 \& 13 \& 14\\ 
            \sor \& \bol \& \bcr\sor \& \sor\bol \& \boli{1} \& \boli{2} \& \scri{2} \& \scri{1} \& \scr\sor \& \sol \& \bcri{1}\bori{1} \& \bcri{1} \& \scr \& \scl  \\ 
        \end{deptext}
        % structural arcs 
        \depedge[edge style={red}]{1}{7}{}
        \depedge[edge style={red}]{3}{8}{}
        \depedge[edge style={red}, edge horizontal padding=-3pt]{4}{9}{}
        \depedge[edge style={red}]{9}{13}{}
        \depedge[edge style={red}, edge horizontal padding=-3pt]{14}{10}{}
        % auxiliar arcs 
        \depedge[dashed]{1}{3}{}
        \depedge[dashed]{7}{2}{}
        \depedge[dashed]{7}{4}{}
        \depedge[dashed]{8}{5}{}
        \depedge[dashed]{9}{6}{}
        \depedge[dashed]{9}{11}{}
        \depedge[dashed]{9}{12}{}
        \depedge[dashed]{11}{14}{}
    \end{dependency}
    \caption{\label{fig:complex-nonprojective-tree}Non-projective extension for a complex graph.}
\end{figure*}

\begin{algorithm*}[h!]
\caption{\label{algo:decodingarbitrary}Extension of Algorithm~\ref{algo:decoding} for arbitrary graphs that can contain crossing arcs. The stack is no longer treated as a strict stack, since we need to be able to access the $i$th element to create crossing arcs. Thus, we assume the following functions: \textsc{Fetch}$(S,\textit{condition},i)$ returns the $i$th element from the top of the stack (the top is the $1$st) satisfying the condition, \textsc{Remove}$(S,\textit{condition},i)$ does the same but removing the element, and \textsc{Put}$(S,\textit{element},i)$ inserts the given element at depth $i$ from the top of the stack (with $i=1$ to push to the top of the stack).}
\begin{small}
\begin{algorithmic}[1]
\Procedure{Decode}{$l_1, \dots, l_n$}
    \State $S \gets$ empty stack ; $A \gets$ empty set
    \For{$i = 1 \to n$}
        \For{symbol$_{\textit{index}}$ in $l_i$} \Comment{\textit{index} = $0$ if the symbol has no index}
            \If{symbol $\in \{{\bol}, {\bor}, {\sol}, {\sor}\}$}
                \State \textsc{Push}($S, (\text{symbol$_{\textit{index}}$}, i)$)
            \ElsIf{symbol $= {\bcr}$}
                \State $(s_x, j) \gets \textsc{Fetch}(S,\textsc{IsSuperbracket},$\textit{index}+1$)$ \Comment{$s \in \{ {\sor},{\sol} \}$}
                \State $A \gets A \cup \{ j \rightarrow i \}$
            \ElsIf{symbol $= {\bcl}$}
                \State $(s_x, j) \gets \textsc{Fetch}(S,\textsc{IsSuperbracket},$\textit{index}+1$)$ \Comment{$s \in \{ {\sor},{\sol} \}$}
                \State $A \gets A \cup \{ j \leftarrow i \}$
            \ElsIf{symbol $={\scr}$ \textbf{or} symbol $={\scl}$}
                \State \Call{CloseSuperbracket}{symbol$_{\textit{index}}$, i, S, A}
            \EndIf
        \EndFor
    \EndFor
\EndProcedure
\newline
\Procedure{CloseSuperbracket}{rbracket$_{\textit{rind}}$, i, S, A}
    \State $\text{depth} \gets 1$
    \State $(\text{lbracket}_{\textit{lind}}, j) \gets \Call{Remove}{S,\textsc{True},\text{depth}}$
    \While{ \textbf{not} ( (lbracket $={\sol}$ \textbf{or} lbracket $={\sor}$) \textbf{and} 
    \textit{rind} = 0 ) }
        \If{ lbracket $={\bol}$ \textbf{or} lbracket $={\bor}$ }
            \If{\textit{lind} > 0}
                \State \Call{Put}{$S$,$(\text{lbracket}_{\textit{lind-1}},j),\text{depth}$}
                \State $\text{depth} \gets \text{depth}+1$
            \ElsIf{lbracket $={\bol}$} $A \gets A \cup \{ j \leftarrow i \}$
            \Else \  $A \gets A \cup \{ j \rightarrow i \}$ \Comment{lbracket $={\bor}$}
            \EndIf
        \Else \Comment{lbracket $={\sol}$ \textbf{or} lbracket $={\sor}$, but $\textit{rind}>0$}
                \State \Call{Put}{$S$,$(\text{lbracket}_{\textit{lind}},j),\text{depth}$}
                \State $\text{depth} \gets \text{depth}+1$
        \EndIf
        \State $(\text{lbracket}_{\textit{lind}}, j) \gets \textsc{Remove}(S,\textsc{True},$\text{depth}$)$
    \EndWhile
    \If{lbracket $={\sol}$ \textbf{and} rbracket $={\scl}$} \Comment{ $\textit{rind} = 0$}
        \State $A \gets A \cup \{ j \leftarrow i \}$
    \ElsIf{lbracket $={\sor}$ \textbf{and} rbracket $={\scr}$}
        \State $A \gets A \cup \{ j \rightarrow i \}$
    \EndIf
\EndProcedure
\end{algorithmic}
\end{small}
\end{algorithm*}

\subsection{Model configuration}\label{ap:model-configuration}
Our neural models consist of a Transformer-based encoder module (XLM-RoBERTa\textsuperscript{L} or XLNet\textsuperscript{L} for English treebanks) stacked with two separate feed-forward networks that predict the sequence of labels and dependency relations using a multi-task cross-entropy loss. The encoder processes the input sentence and returns a contextualized embedding matrix, which is independently fed into each FFN to predict each output label at the token level. For both FFNs, we used the LeakyReLU activation \cite{xu-etal-2015-empirical} and optimized the full architecture with AdamW \cite{loshchilov-etal-2019-decoupled}.

All models were trained for 100 epochs with a constant learning rate $\eta=10^{-5}$, batch size of $300$, and  early stopping on the development set (in terms of UAS) with 20 epochs of patience.

\subsection{Detailed results}\label{subsec:ap-results}
Tables \ref{tab:results-english} to \ref{tab:results-wolof} present a breakdown of results per treebank. Each table includes the Performance on terms of unlabaled (UAS) and labeled (LAS) accuracy and unlabeled (UM) and labeled (LM) exact match on the development and test sets, along with theoretical and empirical coverage.  We compute theoretical coverage based on the arcs recovered when applying the decoding process to the real labels. Empirical coverage is computed similarly but considers the labels and dependency relations that appear in the training set. Thus, empirical coverage sets the maximum performance that an encoding can achieve. Note that for the 4-bit and optimal hierarchical bracketing encodings, theoretical coverage is the same, as both cover only projective graphs and their respective pseudoprojective transformations.

\begin{table}[h]\centering\scriptsize
    \setlength{\tabcolsep}{2pt}
    \renewcommand{\arraystretch}{1.3}
    \begin{tabular}{cc|cccc|cccc|}
        \cline{3-10} 
        \multicolumn{2}{c|}{}& \multicolumn{4}{c|}{\textbf{dev}} & \multicolumn{4}{c|}{\textbf{test}}  \\
        \cline{3-10}
        \multicolumn{2}{c|}{}& \textbf{UAS} & \textbf{LAS} & \textbf{UM} & \textbf{LM} & \textbf{UAS} & \textbf{LAS} & \textbf{UM} & \textbf{LM}\\
        \hline
        \parbox[t]{2mm}{\multirow{4}{*}{\rotatebox[origin=c]{90}{\textbf{th}}}}
        & \textbf{B}\textsubscript{4}/\textbf{O}\SB{P}  & 100 & 100 & 100 & 100 & 100 & 100 & 100 & 100 \\
        & \textbf{B}\SPSB{+}{4}/\textbf{O}\SPSB{+}{P}/\textbf{H}\textsuperscript{+}   & 100 & 100 & 100 & 100 & 100 & 100 & 100 & 100 \\
        & \textbf{B}\textsubscript{7}                   & 100 & 100 & 100 & 100 & 100 & 100 & 100 & 100 \\
        & \textbf{H}\textsuperscript{+}                 & 99.99 & 99.99 & 99.82 & 99.82 & 100 & 100 & 99.96 & 99.96 \\
        \hline 
        \parbox[t]{2mm}{\multirow{7}{*}{\rotatebox[origin=c]{90}{\textbf{emp}}}}
        & \textbf{B}\textsubscript{4}   & 100 & 100 & 100 & 100 & 100 & 100 & 100 & 100 \\
        & \textbf{O}\SB{P}              & 99.99 & 99.99 & 99.82 & 99.82 & 100 & 100 & 99.96 & 99.96 \\
        & \textbf{B}\SPSB{+}{4}         & 100 & 100 & 100 & 100 & 100 & 100 & 100 & 100 \\
        & \textbf{O}\SPSB{+}{P}         &  100 & 100 & 100 & 100 & 100 & 100 & 100 & 100 \\
        & \textbf{B}\textsubscript{7}   & 100 & 100 & 100 & 100 & 100 & 100 & 100 & 100 \\
        & \textbf{O}\SB{NP}             & 100 & 100 & 100 & 100 & 100 & 100 & 100 & 100 \\
        \hline 
        \parbox[t]{2mm}{\multirow{8}{*}{\rotatebox[origin=c]{90}{\textbf{XLNet}}}} 
        & \textbf{B}\textsubscript{4}   & 95.76 & 93.95 & \underline{62.24} & 47.59 & 96.32 & \underline{94.76} & 66.39 & 51.86 \\
        & \textbf{O}\SB{P}              & 95.58 & 93.79 & 61.29 & 47.29 & 96.07 & 94.55 & 65.23 & 51.45 \\
        & \textbf{B}\SPSB{+}{4}         & \underline{95.94} & \underline{94.13} & 62.06 & \underline{48.35} & \underline{96.47} & 94.95 & \underline{66.47} & \underline{52.77} \\
        & \textbf{O}\SPSB{+}{P}         & 94.92 & 93.28 & 59.88 & 47.59 & 95.52 & 93.97 & 63.45 & 49.71 \\
        & \textbf{B}\textsubscript{7}   & \underline{95.67} & 93.76 & \underline{62.12} & \underline{47.12} & \underline{96.14} & \underline{94.61} & \underline{64.98} & \underline{51.99} \\
        & \textbf{O}\SB{NP}             & 95.59 & \underline{93.77} & 61.35 & 47.00 & 95.98 & 94.45 & 65.23 & 51.90 \\
        & \textbf{H}\textsuperscript{+} & 95.67 & 93.83 & 62.18 & 47.53 & 96.30 & 04.80 & 66.43 & 51.95\\
        & \textbf{DM}                   & 96.31 & 94.44 & 62.71 & 48.12 & 96.84 & 95.33 & 65.60 & 51.90 \\
        \hline 
    \end{tabular}
    \caption{\label{tab:results-ptb}Performance on the PTB. Rows are grouped to display the theoretical (\textbf{th}) and empirical (\textbf{emp}) coverage and different encoders (\textbf{XLM}, \textbf{XLNet}).}
\end{table}

\begin{table}[h]\centering\scriptsize
    \setlength{\tabcolsep}{2pt}
    \renewcommand{\arraystretch}{1.3}
    \begin{tabular}{cc|cccc|cccc|}
        \cline{3-10} 
        \multicolumn{2}{c|}{}& \multicolumn{4}{c|}{\textbf{dev}} & \multicolumn{4}{c|}{\textbf{test}}  \\
        \cline{3-10}
        \multicolumn{2}{c|}{}& \textbf{UAS} & \textbf{LAS} & \textbf{UM} & \textbf{LM} & \textbf{UAS} & \textbf{LAS} & \textbf{UM} & \textbf{LM}\\
        \hline
        \parbox[t]{2mm}{\multirow{3}{*}{\rotatebox[origin=c]{90}{\textbf{th}}}}
        & \textbf{B}\textsubscript{4}/\textbf{O}\SB{P}  & 99.80 & 99.80 & 98.50 & 98.50 & 99.83 & 99.83 & 98.94 & 98.94 \\
        & \textbf{B}\SPSB{+}{4}/\textbf{O}\SPSB{+}{P}/\textbf{H}\textsuperscript{+}  &  100 & 100 & 99.95 & 99.95 & 100 & 100 & 100 & 100 \\
        & \textbf{B}\textsubscript{7}                   &  100 & 100 & 100 & 100 & 100 & 100 & 100 & 100 \\
        \hline 
        \parbox[t]{2mm}{\multirow{7}{*}{\rotatebox[origin=c]{90}{\textbf{emp}}}}
        & \textbf{B}\textsubscript{4}   & 99.80 & 99.80 & 98.50 & 98.50 & 99.83 & 99.83 & 98.94 & 98.94 \\
        & \textbf{O}\SB{P}              & 99.84 & 99.84 & 98.35 & 98.35 & 99.88 & 99.88 & 98.89 & 98.89 \\
        & \textbf{B}\SPSB{+}{4}         & 99.99 & 99.99 & 99.70 & 99.70 & 99.99 & 99.99 & 99.90 & 99.90 \\
        & \textbf{O}\SPSB{+}{P}         & 99.99 & 99.99 & 99.70 & 99.70 & 99.99 & 99.99 & 99.90 & 99.90 \\
        & \textbf{B}\textsubscript{7}   & 100 & 100 & 99.95 & 99.95 & 100 & 100 & 100 & 100 \\
        & \textbf{O}\SB{NP}             &  99.98 & 99.98 & 99.90 & 99.90 & 100 & 100 & 100 & 100 \\
        & \textbf{H}\textsuperscript{+} & 99.99 & 99.99 & 99.70 & 99.70 & 99.99 & 99.99 & 99.90 & 99.90 \\
        \hline 
        \parbox[t]{2mm}{\multirow{8}{*}{\rotatebox[origin=c]{90}{\textbf{XLM}}}}
        & \textbf{B}\textsubscript{4}   & 93.51 & 91.27 & 71.31 & 60.42 & 93.01 & 90.76 & 71.83 & 61.53 \\
        & \textbf{O}\SB{P}              & 93.47 & 91.28 & 71.21 & 60.97 & 93.15 & 91.05 & 72.17 & 62.35 \\ 
        & \textbf{B}\SPSB{+}{4}         & 93.76 & 91.62 & 70.76 & 60.72 & 93.27 & 91.01 & 72.36 & 62.16 \\
        & \textbf{O}\SPSB{+}{P}         & 93.65 & 91.37 & 70.96 & 60.42 & 92.96 & 90.54 & 72.65 & 61.48 \\ 
        & \textbf{B}\textsubscript{7}   & 93.64 & 91.39 & 70.81 & 60.02 & 92.99 & 90.69 & 72.22 & 61.92 \\
        & \textbf{O}\SB{NP}             & 93.89 & 91.62 & 71.56 & 60.32 & 93.59 & 91.21 & 72.56 & 61.96 \\ 
        & \textbf{H}\textsuperscript{+} & 94.75 & 92.40 & 73.66 & 62.92 & 94.05 & 91.54 & 74.68 & 63.65 \\
        & \textbf{DM}                   &  95.58 & 93.94 & 75.11 & 66.07 & 95.14 & 93.31 & 76.07 & 66.44 \\
        \hline 
        \parbox[t]{2mm}{\multirow{7}{*}{\rotatebox[origin=c]{90}{\textbf{XLNet}}}}
        & \textbf{B}\textsubscript{4}   & 94.77 & 93.01 & 74.11 & 64.92 & 94.14 & 91.99 & 74.82 & 65.43 \\
        & \textbf{O}\SB{P}              & 94.76 & 92.99 & 74.21 & 64.77 & 94.12 & 92.00 & 75.01 & 64.37 \\
        & \textbf{B}\SPSB{+}{4}         & \underline{95.09} & \underline{93.28} & \underline{74.36} & \underline{65.02} & \underline{94.62} & \underline{92.50} & 75.69 & \underline{66.15} \\
        & \textbf{O}\SPSB{+}{P}         & 94.72 & 92.87 & 74.26 & 64.27 & 94.51 & 92.45 & \underline{75.93} & 65.72 \\
        & \textbf{B}\textsubscript{7}   &  \underline{94.94} & \underline{93.05} & 74.66 & 64.57 & \underline{94.64} & \underline{92.68} & \underline{75.25} & \underline{65.24} \\
        & \textbf{O}\SB{NP}             & 94.69 & 92.86 & \underline{74.86} & \underline{64.57} & 93.87 & 91.69 & 74.29 & 65.00 \\
        & \textbf{H}\textsuperscript{+} & 95.14 & 92.77 & 74.86 & 64.12 & 94.33 & 91.75 & 74.77 & 63.75 \\
        & \textbf{DM}                   & 95.68 & 93.90 & 76.11 & 66.32 & 95.11 & 93.31 & 77.27 & 67.02 \\
        \hline 
    \end{tabular}
    \caption{\label{tab:results-english}Performance on the English EWT.}
\end{table}

\begin{table}[h]\centering\scriptsize
    \setlength{\tabcolsep}{2pt}
    \renewcommand{\arraystretch}{1.3}
    \begin{tabular}{cc|cccc|cccc|}
        \cline{3-10} 
        \multicolumn{2}{c|}{}& \multicolumn{4}{c|}{\textbf{dev}} & \multicolumn{4}{c|}{\textbf{test}}  \\
        \cline{3-10}
        \multicolumn{2}{c|}{}& \textbf{UAS} & \textbf{LAS} & \textbf{UM} & \textbf{LM} & \textbf{UAS} & \textbf{LAS} & \textbf{UM} & \textbf{LM}\\
        \hline
        \parbox[t]{2mm}{\multirow{3}{*}{\rotatebox[origin=c]{90}{\textbf{th}}}}
        & \textbf{B}\textsubscript{4}/\textbf{O}\SB{P} & 86.16 & 86.16 & 40.28 & 40.28 & 85.44 & 85.44 & 47.32 & 47.32 \\
        & \textbf{B}\SPSB{+}{4}/\textbf{O}\SPSB{+}{P}/\textbf{H}\textsuperscript{+} & 99.44 & 99.44 & 90.33 & 90.33 & 99.44 & 99.44 & 92.5 & 92.5 \\
        & \textbf{B}\textsubscript{7}   & 99.33 & 99.33 & 95.34 & 95.34 & 99.24 & 99.24 & 95.25 & 95.25 \\
        \hline 
        \parbox[t]{2mm}{\multirow{7}{*}{\rotatebox[origin=c]{90}{\textbf{emp}}}}
        & \textbf{B}\textsubscript{4}   & 86.16 & 86.16 & 40.28 & 40.28 & 85.44 & 85.43 & 47.32 & 47.24 \\
        & \textbf{O}\textsubscript{P}   & 87.80 & 87.80 & 28.76 & 28.76 & 88.16 & 88.15 & 38.67 & 38.59 \\
        & \textbf{B}\SPSB{+}{4}         &  99.03 & 99.03 & 85.31 & 85.31 & 99.23 & 99.22 & 90.20 & 89.97 \\
        & \textbf{O}\SPSB{+}{4}         & 99.03 & 99.03 & 85.31 & 85.31 & 99.23 & 99.22 & 90.20 & 89.97 \\
        & \textbf{B}\textsubscript{7}   & 99.29 & 99.29 & 95.25 & 95.25 & 99.2 & 99.19 & 95.1 & 94.87 \\
        & \textbf{O}\textsubscript{NP}  & 99.83 & 99.83 & 99.21 & 99.21 & 99.94 & 99.92 & 99.69 & 99.46 \\
        & \textbf{H}\textsuperscript{+} & 98.99 & 98.99 & 85.66 & 85.66 & 99.14 & 99.13 & 89.66 & 89.43 \\
        \hline 
        \parbox[t]{2mm}{\multirow{8}{*}{\rotatebox[origin=c]{90}{\textbf{XLM}}}}
        & \textbf{B}\textsubscript{4}   & 67.19 & 60.04 & 10.55 & 5.36 & 66.74 & 59.64 & 14.70 & 8.81 \\
        & \textbf{O}\SB{P}              & 65.37 & 58.42 & 7.56 & 3.87 & 64.52 & 57.70 & 13.78 & 8.96 \\
        & \textbf{B}\SPSB{+}{4}         & 70.67 & 63.08 & 10.99 & 6.42 & 70.83 & 63.02 & 16.54 & 9.95 \\
        & \textbf{O}\SPSB{+}{P}         &  \underline{70.90} & \underline{63.58} & \underline{12.75} & \underline{7.04} & \underline{71.24} & \underline{63.82} & \underline{18.15} & \underline{11.10} \\
        & \textbf{B}\textsubscript{7}   &  \underline{70.25} & \underline{62.61} & 11.52 & 6.51 & \underline{70.39} & \underline{62.59} & 17.38 & 10.34 \\
        & \textbf{O}\SB{NP}             & 69.37 & 62.57 & \underline{12.84} & \underline{8.00} & 69.51 & 62.46 & \underline{17.53} & \underline{11.10} \\
        & \textbf{H}\textsuperscript{+} & 67.98 & 60.57 & 10.82 & 6.68 & 68.74 & 60.74 & 16.00 & 9.42\\
        & \textbf{DM}                   & 77.37 & 70.25 & 15.92 & 9.32 & 78.36 & 70.79 & 22.89 & 13.40 \\
        \hline 
    \end{tabular}
    \caption{\label{tab:results-greek}Performance on the Ancient-Greek Perseus.}
\end{table}

\begin{table}[h]\centering\scriptsize
    \setlength{\tabcolsep}{2pt}
    \renewcommand{\arraystretch}{1.3}
    \begin{tabular}{cc|cccc|cccc|}
        \cline{3-10} 
        \multicolumn{2}{c|}{}& \multicolumn{4}{c|}{\textbf{dev}} & \multicolumn{4}{c|}{\textbf{test}}  \\
        \cline{3-10}
        \multicolumn{2}{c|}{}& \textbf{UAS} & \textbf{LAS} & \textbf{UM} & \textbf{LM} & \textbf{UAS} & \textbf{LAS} & \textbf{UM} & \textbf{LM}\\
        \hline
        \parbox[t]{2mm}{\multirow{3}{*}{\rotatebox[origin=c]{90}{\textbf{th}}}}
        & \textbf{B}\textsubscript{4}/\textbf{O}\SB{P} & 99.23 & 99.23 & 95.01 & 95.01 & 99.18 & 99.18 & 95.56 & 95.56 \\
        & \textbf{B}\SPSB{+}{4}/\textbf{O}\SPSB{+}{P}/\textbf{H}\textsuperscript{+} &  100 & 100 & 99.93 & 99.93 & 100 & 100 & 99.94 & 99.94 \\
        & \textbf{B}\textsubscript{7}   & 99.99 & 99.99 & 99.93 & 99.93 & 100 & 100 & 99.94 & 99.94 \\
        \hline 
        \parbox[t]{2mm}{\multirow{7}{*}{\rotatebox[origin=c]{90}{\textbf{emp}}}}
        & \textbf{B}\textsubscript{4} & 99.23 & 99.23 & 95.01 & 95.01 & 99.18 & 99.18 & 95.56 & 95.56 \\
        & \textbf{O}\textsubscript{P} & 98.76 & 98.76 & 93.18 & 93.18 & 98.81 & 98.81 & 94.08 & 94.08 \\
        & \textbf{B}\SPSB{+}{4} & 99.97 & 99.97 & 99.34 & 99.34 & 99.99 & 99.99 & 99.74 & 99.74 \\
        & \textbf{O}\SPSB{+}{P} &  99.97 & 99.97 & 99.34 & 99.34 & 99.99 & 99.99 & 99.74 & 99.74 \\
        & \textbf{B}\textsubscript{7} & 99.99 & 99.99 & 99.93 & 99.93 & 99.99 & 99.99 & 99.87 & 99.87 \\
        & \textbf{O}\textsubscript{NP} &100 & 100 & 100 & 100 & 99.96 & 99.96 & 99.87 & 99.87 \\
        & \textbf{H}\textsuperscript{+} & 99.97 & 99.97 & 99.34 & 99.34 & 99.99 & 99.99 & 99.74 & 99.74 \\
        \hline 
        \parbox[t]{2mm}{\multirow{8}{*}{\rotatebox[origin=c]{90}{\textbf{XLM}}}}
        & \textbf{B}\textsubscript{4} & 90.93 & 87.52 & 57.40 & 42.74 & 91.49 & 88.04 & 58.33 & 42.38 \\
        & \textbf{O}\SB{P}            & 91.40 & 87.96 & 60.12 & 45.01 & 91.29 & 88.13 & 59.49 & 44.31 \\
        & \textbf{B}\SPSB{+}{4}       & \underline{92.71} & \underline{89.48} & \underline{63.56} & \underline{47.51} & \underline{92.58} & \underline{89.32} & \underline{61.48} & \underline{45.98} \\
        & \textbf{O}\SPSB{+}{P}       & 92.16 & 88.83 & 61.29 & 46.11 & 92.00 & 88.65 & 60.00 & 44.37 \\
        & \textbf{B}\textsubscript{7} & \underline{92.33} & \underline{89.09} & \underline{62.68} & \underline{47.14} & \underline{92.38} & \underline{89.11} & \underline{61.67} & \underline{45.27} \\
        & \textbf{O}\SB{NP}           & 91.80 & 88.59 & 60.85 & 45.89 & 91.85 & 88.58 & 61.35 & 45.27 \\
        & \textbf{H}\textsuperscript{+} & 92.17 & 88.68 & 62.46 & 46.04 & 92.16 & 88.65 & 60.90 & 44.76 \\
        & \textbf{DM}                 & 94.42 & 81.00 & 67.52 & 0.00 & 94.94 & 81.87 & 68.42 & 0.00 \\
        \hline 
    \end{tabular}
    \caption{\label{tab:results-finnish}Performance on the Finnish TDT.}
\end{table}

\begin{table}[h]\centering\scriptsize
    \setlength{\tabcolsep}{2pt}
    \renewcommand{\arraystretch}{1.3}
    \begin{tabular}{cc|cccc|cccc|}
        \cline{3-10} 
        \multicolumn{2}{c|}{}& \multicolumn{4}{c|}{\textbf{dev}} & \multicolumn{4}{c|}{\textbf{test}}  \\
        \cline{3-10}
        \multicolumn{2}{c|}{}& \textbf{UAS} & \textbf{LAS} & \textbf{UM} & \textbf{LM} & \textbf{UAS} & \textbf{LAS} & \textbf{UM} & \textbf{LM}\\
        \hline
        \parbox[t]{2mm}{\multirow{3}{*}{\rotatebox[origin=c]{90}{\textbf{th}}}}
        & \textbf{B}\textsubscript{4}/\textbf{O}\SB{P} & 99.46 & 99.46 & 96.27 & 96.27 & 99.35 & 99.35 & 95.91 & 95.91 \\
        & \textbf{B}\SPSB{+}{4}/\textbf{O}\SPSB{+}{P}/\textbf{H}\textsuperscript{+}  & 99.98 & 99.98 & 99.59 & 99.59 & 99.99 & 99.99 & 99.76 & 99.76 \\
        & \textbf{B}\textsubscript{7}   & 100 & 100 & 100 & 100 & 100 & 100 & 100 & 100 \\
        \hline 
        \parbox[t]{2mm}{\multirow{7}{*}{\rotatebox[origin=c]{90}{\textbf{emp}}}}
        & \textbf{B}\textsubscript{4} & 99.46 & 99.46 & 96.27 & 96.27 & 99.35 & 99.35 & 95.91 & 95.91 \\
        & \textbf{O}\textsubscript{P} & 99.69 & 99.69 & 96.41 & 96.41 & 99.67 & 99.67 & 96.15 & 96.15 \\
        & \textbf{B}\SPSB{+}{4} &99.98 & 99.98 & 99.46 & 99.46 & 99.98 & 99.98 & 99.28 & 99.28 \\
        & \textbf{O}\SPSB{+}{P} &99.98 & 99.98 & 99.46 & 99.46 & 99.98 & 99.98 & 99.28 & 99.28 \\
        & \textbf{B}\textsubscript{7} & 100 & 100 & 100 & 100 & 100 & 100 & 100 & 100 \\
        & \textbf{O}\textsubscript{NP} & 100 & 100 & 100 & 100 & 100 & 100 & 100 & 100 \\
        & \textbf{H}\textsuperscript{+} & 99.98 & 99.98 & 99.46 & 99.46 & 99.98 & 99.98 & 99.28 & 99.28 \\
        \hline 
        \parbox[t]{2mm}{\multirow{8}{*}{\rotatebox[origin=c]{90}{\textbf{XLM}}}}
        & \textbf{B}\textsubscript{4}  & 96.19 & 94.09 & 58.74 & 43.36 & 94.65 & 91.90 & 50.24 & 34.86 \\
        & \textbf{O}\textsubscript{4} & 96.13 & 94.08 & 58.67 & 42.82 & 94.33 & 91.40 & \underline{51.68} & 35.58 \\
        & \textbf{B}\SPSB{+}{4} & \underline{96.52} & \underline{94.42} & \underline{60.09} & 44.17 & 94.70 & \underline{92.19} & 49.28 & \underline{36.30} \\
        & \textbf{O}\SPSB{+}{P} & 96.44 & 94.39 & 59.82 & \underline{44.38} & \underline{94.81} & 92.06 & 49.04 & 35.34 \\
        & \textbf{B}\textsubscript{7} & \underline{96.46} & \underline{94.42} & 59.55 & \underline{44.58} & \underline{94.45} & \underline{91.82} & 48.32 & 35.10 \\
        & \textbf{O}\textsubscript{NP} & 96.37 & 94.23 & \underline{59.96} & 44.04 & 94.14 & 91.59 & \underline{48.80} & \underline{37.02} \\
        & \textbf{H}\textsuperscript{+} & 95.78 & 93.75 & 58.13 & 43.29 & 94.53 & 91.50 & 50.48 & 33.41 \\ 
        & \textbf{DM} & 97.33 & 95.46 & 62.87 & 46.75 & 96.03 & 93.57 & 52.16 & 39.18 \\
        \hline 
    \end{tabular}
    \caption{\label{tab:results-french}Performance on the French GSD.}
\end{table}

\begin{table}[h]\centering\scriptsize
    \setlength{\tabcolsep}{2pt}
    \renewcommand{\arraystretch}{1.3}
    \begin{tabular}{cc|cccc|cccc|}
        \cline{3-10} 
        \multicolumn{2}{c|}{}& \multicolumn{4}{c|}{\textbf{dev}} & \multicolumn{4}{c|}{\textbf{test}}  \\
        \cline{3-10}
        \multicolumn{2}{c|}{}& \textbf{UAS} & \textbf{LAS} & \textbf{UM} & \textbf{LM} & \textbf{UAS} & \textbf{LAS} & \textbf{UM} & \textbf{LM}\\
        \hline
        \parbox[t]{2mm}{\multirow{3}{*}{\rotatebox[origin=c]{90}{\textbf{th}}}}
        & \textbf{B}\textsubscript{4}/\textbf{O}\SB{P} & 99.98 & 99.98 & 99.79 & 99.79 & 100 & 100 & 100 & 100 \\
        & \textbf{B}\SPSB{+}{4}/\textbf{O}\SPSB{+}{P}/\textbf{H}\textsuperscript{+}  & 100 & 100 & 100 & 100 & 100 & 100 & 100 & 100 \\
        & \textbf{B}\textsubscript{7}   & 100 & 100 & 100 & 100 & 100 & 100 & 100 & 100 \\
        \hline 
        \parbox[t]{2mm}{\multirow{7}{*}{\rotatebox[origin=c]{90}{\textbf{emp}}}}
        & \textbf{B}\textsubscript{4} & 99.98 & 99.98 & 99.79 & 99.79 & 100 & 100 & 100 & 100 \\
        & \textbf{O}\textsubscript{P} & 99.95 & 99.95 & 99.38 & 99.38 & 99.90 & 99.90 & 99.59 & 99.59 \\
        & \textbf{B}\SPSB{+}{4} & 100 & 100 & 100 & 100 & 100 & 100 & 100 & 100 \\
        & \textbf{O}\SPSB{+}{4} & 100 & 100 & 100 & 100 & 100 & 100 & 100 & 100 \\
        & \textbf{B}\textsubscript{7} &  100 & 100 & 100 & 100 & 100 & 100 & 100 & 100 \\
        & \textbf{H}\textsuperscript{+} & 100 & 100 & 100 & 100 & 100 & 100 & 100 & 100 \\
        & \textbf{O}\textsubscript{NP} & 100 & 100 & 100 & 100 & 100 & 100 & 100 & 100 \\
        \hline 
        \parbox[t]{2mm}{\multirow{8}{*}{\rotatebox[origin=c]{90}{\textbf{XLM}}}}
        & \textbf{B}\textsubscript{4} & \underline{93.83} & \underline{91.02} & 48.14 & \underline{35.33} & \underline{92.78} & \underline{90.14} & \underline{42.77} & \underline{30.35} \\
        & \textbf{O}\textsubscript{P} & 93.47 & 90.39 & 46.49 & 32.85 & 92.06 & 89.51 & 42.57 & 30.35 \\
        & \textbf{B}\SPSB{+}{4} & 93.64 & 90.86 & 47.11 & 33.47 & 92.21 & 89.51 & 40.33 & 28.92 \\
        & \textbf{O}\SPSB{+}{4}& 93.43 & 90.47 & \underline{48.35} & 34.09 & 91.79 & 89.21 & 40.33 & 27.09 \\
        & \textbf{B}\textsubscript{7} & 93.59 & \underline{91.03} & 46.69 & \underline{36.16} & \underline{92.33} & 89.59 & \underline{42.77} & \underline{28.92} \\
        & \textbf{O}\textsubscript{NP} & \underline{93.88} & 90.39 & \underline{50.21} & 33.88 & 92.32 & \underline{89.61} & 42.16 & 28.92 \\
        & \textbf{H}\textsuperscript{+} & 92.37 & 88.64 & 45.04 & 27.89 & 91.25 & 87.86 & 38.70 & 25.46\\
        & \textbf{DM} & 95.87 & 93.42 & 54.55 & 40.70 & 94.42 & 92.02 & 48.47 & 33.81 \\
        \hline 
    \end{tabular}
    \caption{\label{tab:results-hebrew}Performance on the Hebrew HTB.}
\end{table}

\begin{table}[h]\centering\scriptsize
    \setlength{\tabcolsep}{2pt}
    \renewcommand{\arraystretch}{1.3}
    \begin{tabular}{cc|cccc|cccc|}
        \cline{3-10} 
        \multicolumn{2}{c|}{}& \multicolumn{4}{c|}{\textbf{dev}} & \multicolumn{4}{c|}{\textbf{test}}  \\
        \cline{3-10}
        \multicolumn{2}{c|}{}& \textbf{UAS} & \textbf{LAS} & \textbf{UM} & \textbf{LM} & \textbf{UAS} & \textbf{LAS} & \textbf{UM} & \textbf{LM}\\
        \hline
        \parbox[t]{2mm}{\multirow{3}{*}{\rotatebox[origin=c]{90}{\textbf{th}}}}
        & \textbf{B}\textsubscript{4}/\textbf{O}\SB{P} & 99.41 & 99.41 & 95.16 & 95.16 & 99.44 & 99.44 & 95.84 & 95.84 \\
        & \textbf{B}\SPSB{+}{4}/\textbf{O}\SPSB{+}{P}/\textbf{H}\textsuperscript{+} &99.98 & 99.98 & 99.31 & 99.31 & 99.96 & 99.96 & 99.17 & 99.17 \\
        & \textbf{B}\textsubscript{7}   & 100 & 100 & 100 & 100 & 100 & 100 & 100 & 100 \\
        \hline 
        \parbox[t]{2mm}{\multirow{7}{*}{\rotatebox[origin=c]{90}{\textbf{emp}}}}
        & \textbf{B}\textsubscript{4} &  99.41 & 99.41 & 95.16 & 95.16 & 99.44 & 99.43 & 95.84 & 95.84 \\
        & \textbf{O}\textsubscript{P} &99.39 & 99.39 & 94.47 & 94.47 & 99.23 & 99.23 & 94.34 & 94.34 \\
        & \textbf{B}\SPSB{+}{4} & 99.90 & 99.90 & 97.41 & 97.41 & 99.81 & 99.81 & 96.51 & 96.51 \\
        & \textbf{O}\SPSB{+}{P} & 99.90 & 99.90 & 97.41 & 97.41 & 99.81 & 99.81 & 96.51 & 96.51 \\
        & \textbf{B}\textsubscript{7} & 99.97 & 99.97 & 99.83 & 99.83 & 100 & 100 & 100 & 99.83 \\
        & \textbf{O}\textsubscript{NP} &99.99 & 99.99 & 99.65 & 99.65 & 99.98 & 99.98 & 99.83 & 99.67 \\
        & \textbf{H}\textsuperscript{+} &99.90 & 99.90 & 97.41 & 97.41 & 99.81 & 99.81 & 96.51 & 96.51 \\
        \hline 
        \parbox[t]{2mm}{\multirow{8}{*}{\rotatebox[origin=c]{90}{\textbf{XLM}}}}
        & \textbf{B}\textsubscript{4} & 91.83 & 88.95 & \underline{44.39} & \underline{32.99} & 91.76 & 87.99 & 45.42 & \underline{31.11} \\
        & \textbf{O}\textsubscript{P} & 91.15 & 88.17 & 41.28 & 30.92 & 91.49 & 87.82 & 46.92 & 30.62 \\
        & \textbf{B}\SPSB{+}{4} & \underline{92.12} & \underline{89.27} & 41.97 & 31.61 & \underline{92.08} & 88.09 & 45.59 & 28.79 \\
        & \textbf{O}\SPSB{+}{P} & 91.29 & 88.58 & 43.01 & 32.30 & 91.85 & \underline{88.18} & \underline{46.92} & 30.95 \\
        & \textbf{B}\textsubscript{7} & \underline{91.66} & \underline{88.73} & \underline{41.45} & \underline{30.40} & \underline{92.14} & \underline{88.40} & \underline{46.26} & \underline{30.95} \\
        & \textbf{O}\textsubscript{NP} & 90.85 & 88.03 & 41.11 & 30.40 & 90.85 & 87.06 & 44.76 & 28.79 \\
        & \textbf{H}\textsuperscript{+} & 90.13 & 87.45 & 40.07 & 30.92 & 91.63 & 87.61 & 45.59 & 30.62 \\
        & \textbf{DM} & 93.94 & 91.50 & 47.67 & 37.48 & 93.99 & 90.60 & 53.74 & 36.44 \\
        \hline 
    \end{tabular}
    \caption{\label{tab:results-russian}Performance on the Russian GSD.}
\end{table}

\begin{table}[h]\centering\scriptsize
    \setlength{\tabcolsep}{2pt}
    \renewcommand{\arraystretch}{1.3}
    \begin{tabular}{cc|cccc|cccc|}
        \cline{3-10} 
        \multicolumn{2}{c|}{}& \multicolumn{4}{c|}{\textbf{dev}} & \multicolumn{4}{c|}{\textbf{test}}  \\
        \cline{3-10}
        \multicolumn{2}{c|}{}& \textbf{UAS} & \textbf{LAS} & \textbf{UM} & \textbf{LM} & \textbf{UAS} & \textbf{LAS} & \textbf{UM} & \textbf{LM}\\
        \hline
        \parbox[t]{2mm}{\multirow{3}{*}{\rotatebox[origin=c]{90}{\textbf{th}}}}
        & \textbf{B}\textsubscript{4}/\textbf{O}\SB{P} & 100 & 100 & 100 & 100 & 99.87 & 99.87 & 98.33 & 98.33 \\
        & \textbf{B}\SPSB{+}{4}/\textbf{O}\SPSB{+}{P}/\textbf{H}\textsuperscript{+}  &  100 & 100 & 100 & 100 & 100 & 100 & 100 & 100 \\
        & \textbf{B}\textsubscript{7}   & 100 & 100 & 100 & 100 & 100 & 100 & 100 & 100 \\
        \hline 
        \parbox[t]{2mm}{\multirow{7}{*}{\rotatebox[origin=c]{90}{\textbf{emp}}}}
        & \textbf{B}\textsubscript{4} & 100 & 99.96 & 100 & 98.75 & 99.87 & 99.87 & 98.33 & 98.33 \\
        & \textbf{O}\textsubscript{P} & 100 & 99.96 & 100 & 98.75 & 99.82 & 99.82 & 97.50 & 97.50 \\
        & \textbf{B}\SPSB{+}{4} & 100 & 99.96 & 100 & 98.75 & 99.84 & 99.84 & 97.50 & 97.50 \\
        & \textbf{O}\SPSB{+}{P} & 100 & 99.96 & 100 & 98.75 & 99.84 & 99.84 & 97.50 & 97.50 \\
        & \textbf{B}\textsubscript{7} & 100 & 99.96 & 100 & 98.75 & 99.87 & 99.87 & 98.33 & 98.33 \\
        & \textbf{O}\textsubscript{NP} & 100 & 99.96 & 100 & 98.75 & 99.93 & 99.93 & 99.17 & 99.17 \\
        & \textbf{H}\textsuperscript{+} & 100 & 99.96 & 100 & 98.75 & 99.84 & 99.84 & 97.50 & 97.50 \\
        \hline 
        \parbox[t]{2mm}{\multirow{8}{*}{\rotatebox[origin=c]{90}{\textbf{XLM}}}}
        & \textbf{B}\textsubscript{4} & 78.72 & 69.10 & 17.50 & 6.25 & 76.07 & 65.54 & 14.17 & 2.50 \\
        & \textbf{O}\textsubscript{P} & 79.30 & 69.03 & \underline{23.75} & 8.75 & 75.82 & 64.80 & \underline{16.67} & \underline{5.00} \\
        & \textbf{B}\SPSB{+}{4} & \underline{79.33} & 69.64 & 20.00 & 6.25 & \underline{77.11} & \underline{66.57} & 15.83 & 2.50 \\
        & \textbf{O}\SPSB{+}{P} & 78.52 & \underline{69.73} & 16.25 & \underline{10.00} & 76.69 & 66.06 & 15.83 & 1.67 \\
        & \textbf{B}\textsubscript{7} & \underline{79.06} & \underline{68.97} & \underline{18.75} & \underline{8.75} & \underline{78.02} & \underline{68.12} & \underline{19.17} & \underline{5.00} \\
        & \textbf{O}\textsubscript{NP} & 77.76 & 67.37 & 15.00 & 5.00 & 77.09 & 65.26 & 19.17 & 3.33 \\
        & \textbf{H}\textsuperscript{+} & 78.22 & 67.23 & 16.25 & 5.00 &  76.26 & 65.13 & 17.50 & 3.33 \\
        & \textbf{DM} & 82.85 & 73.47 & 25.00 & 10.00 & 81.48 & 70.68 & 24.17 & 6.67 \\
        \hline 
    \end{tabular}
    \caption{\label{tab:results-tamil}Performance on the Tamil TTB.}
\end{table}

\begin{table}[h]\centering\scriptsize
    \setlength{\tabcolsep}{2pt}
    \renewcommand{\arraystretch}{1.3}
    \begin{tabular}{cc|cccc|cccc|}
        \cline{3-10} 
        \multicolumn{2}{c|}{}& \multicolumn{4}{c|}{\textbf{dev}} & \multicolumn{4}{c|}{\textbf{test}}  \\
        \cline{3-10}
        \multicolumn{2}{c|}{}& \textbf{UAS} & \textbf{LAS} & \textbf{UM} & \textbf{LM} & \textbf{UAS} & \textbf{LAS} & \textbf{UM} & \textbf{LM}\\
        \hline
        \parbox[t]{2mm}{\multirow{3}{*}{\rotatebox[origin=c]{90}{\textbf{th}}}}
        & \textbf{B}\textsubscript{4}/\textbf{O}\SB{P} & 99.01 & 99.01 & 92.33 & 92.33 & 98.66 & 98.66 & 92.33 & 92.33 \\
        & \textbf{B}\SPSB{+}{4}/\textbf{O}\SPSB{+}{P}/\textbf{H}\textsuperscript{+}  & 99.98 & 99.98 & 99.44 & 99.44 & 99.96 & 99.96 & 99.33 & 99.33 \\
        & \textbf{B}\textsubscript{7}   & 100 & 100 & 100 & 100 & 100 & 100 & 100 & 100 \\
        \hline 
        \parbox[t]{2mm}{\multirow{7}{*}{\rotatebox[origin=c]{90}{\textbf{emp}}}}
        & \textbf{B}\textsubscript{4} & 98.83 & 98.75 & 93.11 & 92.11 & 98.50 & 97.96 & 93.44 & 87.44 \\
        & \textbf{O}\textsubscript{P} & 99.01 & 98.94 & 92.33 & 91.33 & 98.66 & 98.13 & 92.33 & 86.44 \\
        & \textbf{B}\SPSB{+}{4} & 99.83 & 99.76 & 97.67 & 96.56 & 99.81 & 99.25 & 97.33 & 91.22 \\
        & \textbf{O}\SPSB{+}{P} & 99.83 & 99.76 & 97.67 & 96.56 & 99.81 & 99.25 & 97.33 & 91.22 \\
        & \textbf{B}\textsubscript{7} & 98.83 & 98.75 & 93.11 & 92.11 & 98.50 & 97.96 & 93.44 & 87.44 \\
        & \textbf{O}\textsubscript{NP} & 99.88 & 99.81 & 99.22 & 98.00 & 99.68 & 99.13 & 99.00 & 92.78 \\
        & \textbf{H}\textsuperscript{+} & 99.82 & 99.75 & 97.44 & 96.33 & 99.8 & 99.25 & 97.33 & 91.22 \\
        \hline 
        \parbox[t]{2mm}{\multirow{8}{*}{\rotatebox[origin=c]{90}{\textbf{XLM}}}}
        & \textbf{B}\textsubscript{4} & 81.50 & \underline{68.22} & \underline{31.67} & \underline{11.44} & \underline{80.03} & 66.62 & \underline{30.00} & 11.67 \\
        & \textbf{O}\textsubscript{P} & 80.26 & 67.02 & 29.89 & 10.67 & 79.37 & 66.27 & 29.44 & 11.89 \\
        & \textbf{B}\SPSB{+}{4} & \underline{81.50} & 68.10 & 30.78 & 10.78 & 79.70 & 66.78 & 29.00 & 11.22 \\
        & \textbf{O}\SPSB{+}{P} & 80.83 & 67.78 & 30.22 & 10.33 & 79.55 & \underline{66.88} & 28.78 & \underline{12.33} \\
        & \textbf{B}\textsubscript{7} & \underline{81.03} & \underline{67.49} & 30.67 & \underline{10.89} & \underline{80.09} & \underline{66.47} & \underline{30.11} & 10.33 \\
        & \textbf{O}\textsubscript{NP} & 80.15 & 66.23 & \underline{30.78} & 9.11 & 78.70 & 65.17 & 29.33 & \underline{11.22} \\
        & \textbf{H}\textsuperscript{+} & 81.65 & 67.37 & 32.33 & 10.11 & 79.52 & 65.56 & 29.33 & 10.00 \\
        & \textbf{DM} & 85.08 & 72.84 & 38.44 & 15.22 & 84.16 & 71.39 & 38.44 & 14.56 \\
        \hline 
    \end{tabular}
    \caption{\label{tab:results-uyghur}Performance on the Uyghur UDT.}
\end{table}

\begin{table}[h]\centering\scriptsize
    \setlength{\tabcolsep}{2pt}
    \renewcommand{\arraystretch}{1.3}
    \begin{tabular}{cc|cccc|cccc|}
        \cline{3-10} 
        \multicolumn{2}{c|}{}& \multicolumn{4}{c|}{\textbf{dev}} & \multicolumn{4}{c|}{\textbf{test}}  \\
        \cline{3-10}
        \multicolumn{2}{c|}{}& \textbf{UAS} & \textbf{LAS} & \textbf{UM} & \textbf{LM} & \textbf{UAS} & \textbf{LAS} & \textbf{UM} & \textbf{LM}\\
        \hline
        \parbox[t]{2mm}{\multirow{3}{*}{\rotatebox[origin=c]{90}{\textbf{th}}}}
        & \textbf{B}\textsubscript{4}/\textbf{O}\SB{P} & 99.68 & 99.68 & 96.44 & 96.44 & 99.64 & 99.64 & 97.02 & 97.02 \\
        & \textbf{B}\SPSB{+}{4}/\textbf{O}\SPSB{+}{P}/\textbf{H}\textsuperscript{+}  & 100 & 100 & 100 & 100 & 99.99 & 99.99 & 99.79 & 99.79 \\
        & \textbf{B}\textsubscript{7}   & 100 & 100 & 100 & 100 & 100 & 100 & 100 & 100 \\
        \hline 
        \parbox[t]{2mm}{\multirow{7}{*}{\rotatebox[origin=c]{90}{\textbf{emp}}}}
        & \textbf{B}\textsubscript{4} & 99.71 & 99.69 & 96.88 & 96.44 & 99.58 & 99.57 & 96.81 & 96.60 \\
        & \textbf{O}\textsubscript{P} & 99.68 & 99.67 & 96.44 & 95.99 & 99.64 & 99.63 & 97.02 & 96.81 \\
        & \textbf{B}\SPSB{+}{4} & 99.93 & 99.92 & 98.22 & 97.77 & 99.93 & 99.92 & 98.72 & 98.51 \\
        & \textbf{O}\SPSB{+}{P} & 99.93 & 99.92 & 98.22 & 97.77 & 99.93 & 99.92 & 98.72 & 98.51 \\
        & \textbf{B}\textsubscript{7} & 99.71 & 99.69 & 96.88 & 96.44 & 99.58 & 99.57 & 96.81 & 96.60 \\
        & \textbf{O}\textsubscript{NP} & 99.92 & 99.91 & 99.55 & 99.11 & 100 & 99.99 & 100 & 99.79 \\
        & \textbf{H}\textsuperscript{+} &  99.93 & 99.92 & 98.0 & 97.55 & 99.90 & 99.89 & 98.3 & 98.09 \\
        \hline 
        \parbox[t]{2mm}{\multirow{8}{*}{\rotatebox[origin=c]{90}{\textbf{XLM}}}}
        & \textbf{B}\textsubscript{4} & \underline{81.15} & \underline{73.82} & \underline{15.37} & \underline{6.90} & \underline{80.82} & \underline{73.67} & 18.30 & \underline{9.15} \\
        & \textbf{O}\textsubscript{P} & 79.30 & 71.03 & 14.03 & 6.46 & 78.33 & 70.25 & 18.09 & 7.66 \\
        & \textbf{B}\SPSB{+}{4} & 80.43 & 72.78 & 13.81 & 5.12 & 80.37 & 73.02 & \underline{19.15} & 8.30 \\
        & \textbf{O}\SPSB{+}{P} & 79.35 & 71.46 & 13.36 & 5.79 & 78.81 & 71.13 & 18.72 & 8.30 \\
        & \textbf{B}\textsubscript{7} & \underline{80.54} & \underline{72.94} & 15.14 & \underline{6.68} & \underline{80.62} & \underline{72.84} & \underline{21.28} & \underline{11.49} \\
        & \textbf{O}\textsubscript{NP} & 80.47 & 72.58 & \underline{16.70} & 6.68 & 79.90 & 72.37 & 18.94 & 8.72 \\
        & \textbf{H}\textsuperscript{+} & 77.08 & 68.13 & 13.14 & 4.68 & 75.65 & 67.23 & 15.11 & 5.11 \\
        & \textbf{DM} & 80.70 & 72.37 & 16.70 & 6.90 & 81.34 & 73.56 & 20.43 & 10.00 \\
        \hline 
    \end{tabular}
    \caption{\label{tab:results-wolof}Performance on the Wolof WTB.}
\end{table}

\end{document}